\pdfoutput=1
\documentclass[11pt]{article}

\usepackage[final]{acl}

\usepackage{times}
\usepackage{latexsym}
\usepackage{booktabs}
\usepackage{enumitem}
\usepackage{algorithm}
\usepackage{algpseudocode}
\usepackage{tcolorbox}
\usepackage{titlesec}
\tcbuselibrary{breakable}
\tcbuselibrary{listings}
\usepackage{caption}  
\usepackage{graphicx}
\usepackage{subcaption}
\usepackage{multirow}
\usepackage[T1]{fontenc}
\usepackage[utf8]{inputenc}

\usepackage{microtype}

\usepackage{inconsolata}

\usepackage{graphicx}

\newcommand{\InterChart}{\textsc{InterChart}}
\title{\InterChart: Benchmarking Visual Reasoning Across Decomposed and Distributed Chart Information}

\newcommand{\asulogo}{\includegraphics[height=8pt]{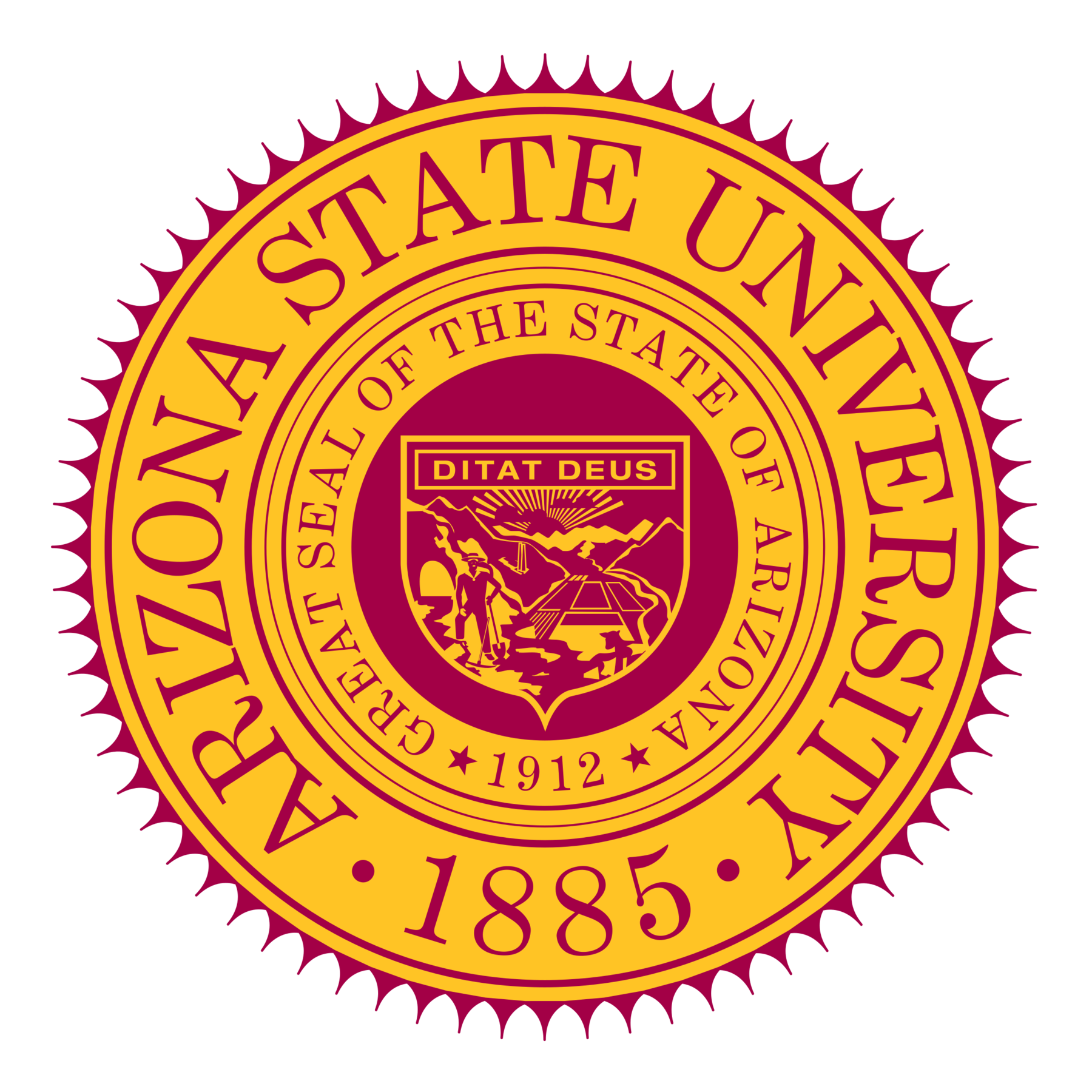}}
\newcommand{\iithlogo}{\includegraphics[height=8pt]{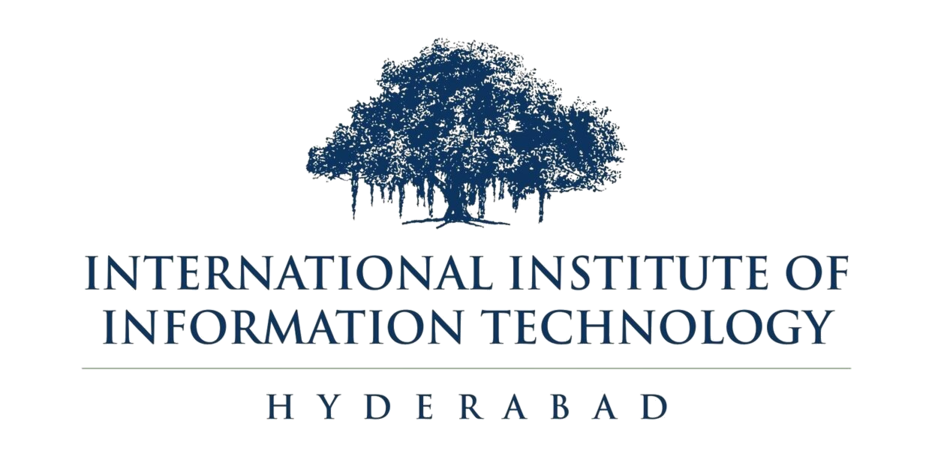}} 
\newcommand{\mercerlogo}{\includegraphics[height=10pt]{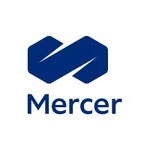}}
\newcommand{\upenlogo}{\includegraphics[height=8pt]{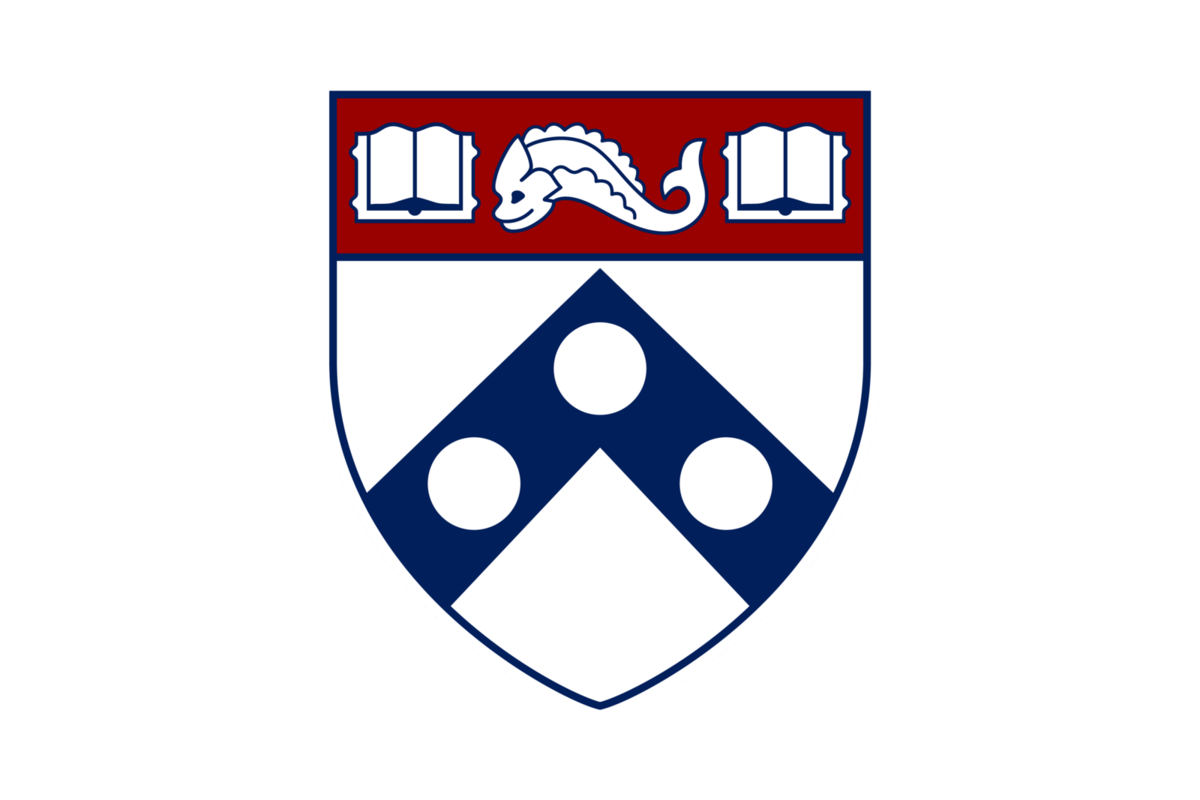}}

\author{
\textbf{Anirudh Iyengar Kaniyar Narayana Iyengar}\asulogo\textbf{$^{*}$}, 
\textbf{Srija Mukhopadhyay}\iithlogo\textbf{$^{*}$}, \\
\textbf{Adnan Qidwai}\iithlogo\textbf{$^{*}$},
\textbf{Shubhankar Singh}\mercerlogo, 
\textbf{Dan Roth}\upenlogo, 
\textbf{Vivek Gupta}\asulogo
\\[2pt]
{\asulogo Arizona State University 
\iithlogo IIIT, Hyderabad 
\mercerlogo Mercer Mettl 
\upenlogo University of Pennsylvania}
\\[0pt]
\footnotesize \texttt{akaniyar@asu.edu, srija.mukhopadhyay@research.iiit.ac.in, adnan.qidwai@students.iiit.ac.in,}\\[-3pt]
\footnotesize \texttt{ Shubhankar.singh@mercer.com, danroth@seas.upenn.edu, vgupt140@asu.edu}
}

\begin{document}
\maketitle
\begin{abstract}
We introduce \textbf{\InterChart}, a diagnostic benchmark that evaluates how well vision-language models (VLMs) reason across multiple related charts, a task central to real-world applications such as scientific reporting, financial analysis, and public policy dashboards. Unlike prior benchmarks focusing on isolated, visually uniform charts, \textbf{\InterChart{}} challenges models with diverse question types ranging from entity inference and trend correlation to numerical estimation and abstract multi-step reasoning grounded in 2-3 thematically or structurally related charts. We organize the benchmark into three tiers of increasing difficulty: (1) factual reasoning over individual charts, (2) integrative analysis across synthetically aligned chart sets, and (3) semantic inference over visually complex, real-world chart pairs. Our evaluation of state-of-the-art open- and closed-source VLMs reveals consistent and steep accuracy declines as chart complexity increases. We find that models perform better when we decompose multi-entity charts into simpler visual units, underscoring their struggles with cross-chart integration. By exposing these systematic limitations, \InterChart{} provides a rigorous framework for advancing multimodal reasoning in complex, multi-visual environments.
\end{abstract}

\let\thefootnote\relax\footnote{$^{*}$These authors contributed equally to this work}
\section{Introduction}
Real-world settings such as scientific publications, business reports, and journalism dashboards rarely communicate data through a single chart. Instead, insight often emerges from comparing or synthesizing information across multiple visualizations. These charts may differ in type, styling, or even semantic framing, yet they jointly convey trends, correlations, and complex relationships. For humans, reasoning across such heterogeneous visual inputs is intuitive. However, vision-language models (VLMs) continue to face significant challenges when required to integrate information across visually heterogeneous chart collections.

While recent VLMs have shown strong performance on single-chart visual question answering (VQA) tasks~\cite{masry2022chartqa, methani2020plotqa}, they perform inconsistently to aggregate information across multiple charts. Existing benchmarks~\cite{li2023scigraphqa, kantharaj2022chart} have begun exploring multi-chart reasoning, but they often rely on simplified scenarios, synthetic data, static chart styles, or limited visual variation. Consequently, these datasets fail to capture key challenges in real-world chart reasoning: visual inconsistency, semantic misalignment, temporal discontinuity, and multi-step aggregation. Moreover, their evaluation metrics typically depend on string matching, which inadequately reflects semantic understanding.

We introduce \textbf{\InterChart{}}, a diagnostic benchmark designed to probe how well VLMs can reason across multiple charts with increasing levels of complexity. Unlike prior datasets, \InterChart{} spans both synthetic and real-world charts, and introduces a structured tiering system to evaluate performance under controlled and unconstrained conditions. It targets a range of reasoning abilities-from simple fact extraction to multi-step, cross-domain inference-allowing researchers to disentangle visual parsing errors from reasoning failures.

\InterChart{} is organized into three structured subsets, each designed to isolate distinct reasoning challenges rather than to establish a predictive hierarchy. The first tier, \emph{DECAF} (Decomposed Elementary Charts with Answerable Facts), evaluates atomic fact retrieval and localized comparisons within visually simplified, decomposed charts. The second tier, \emph{SPECTRA} (Synthetic Plots for Event-based Correlated Trend Reasoning and Analysis), probes correlated trend reasoning across synthetic chart pairs that share axes and stylistic variations, testing a model’s ability to align related quantities and interpret event-based trends. The third and most advanced tier, \emph{STORM} (Sequential Temporal Reasoning Over Real-world Multi-domain charts), examines semantic abstraction and temporal alignment across visually and thematically diverse real-world chart pairs. Collectively, these subsets serve a diagnostic purpose revealing model-specific failure modes tied to visual complexity, semantic drift, and temporal aggregation rather than implying transferable performance or ranking consistency across tiers.

To ensure reliable assessment, we propose a novel LLM-assisted evaluation pipeline. Instead of relying solely on an exact string match, we employ multiple LLMs as semantic judges and aggregate their decisions through majority voting. It enables evaluators to assess paraphrased answers, numeric approximations, and equivalent units flexibly, producing more robust performance estimates.

\noindent We summarize our contributions as follows:
\begin{enumerate}
\setlength\itemsep{0em}
\item We present \textbf{\InterChart{}}, the first multi-tier benchmark for multi-chart VQA, spanning decomposed, synthetic, and real-world chart contexts.
\item We design structured reasoning tasks to benchmark on various closed and open-source VLMs across three visual tiers, capturing localized and cross-visual dependencies, including trend correlation and temporal abstraction.
\item We propose an LLM-assisted semantic evaluation framework that improves alignment with human judgment and enables fine-grained error analysis.
\end{enumerate}

The dataset and resources are publicly available at \url{https://coral-lab-asu.github.io/interchart/}.

\begin{figure*}[t]
\centering
\includegraphics[width=\linewidth]{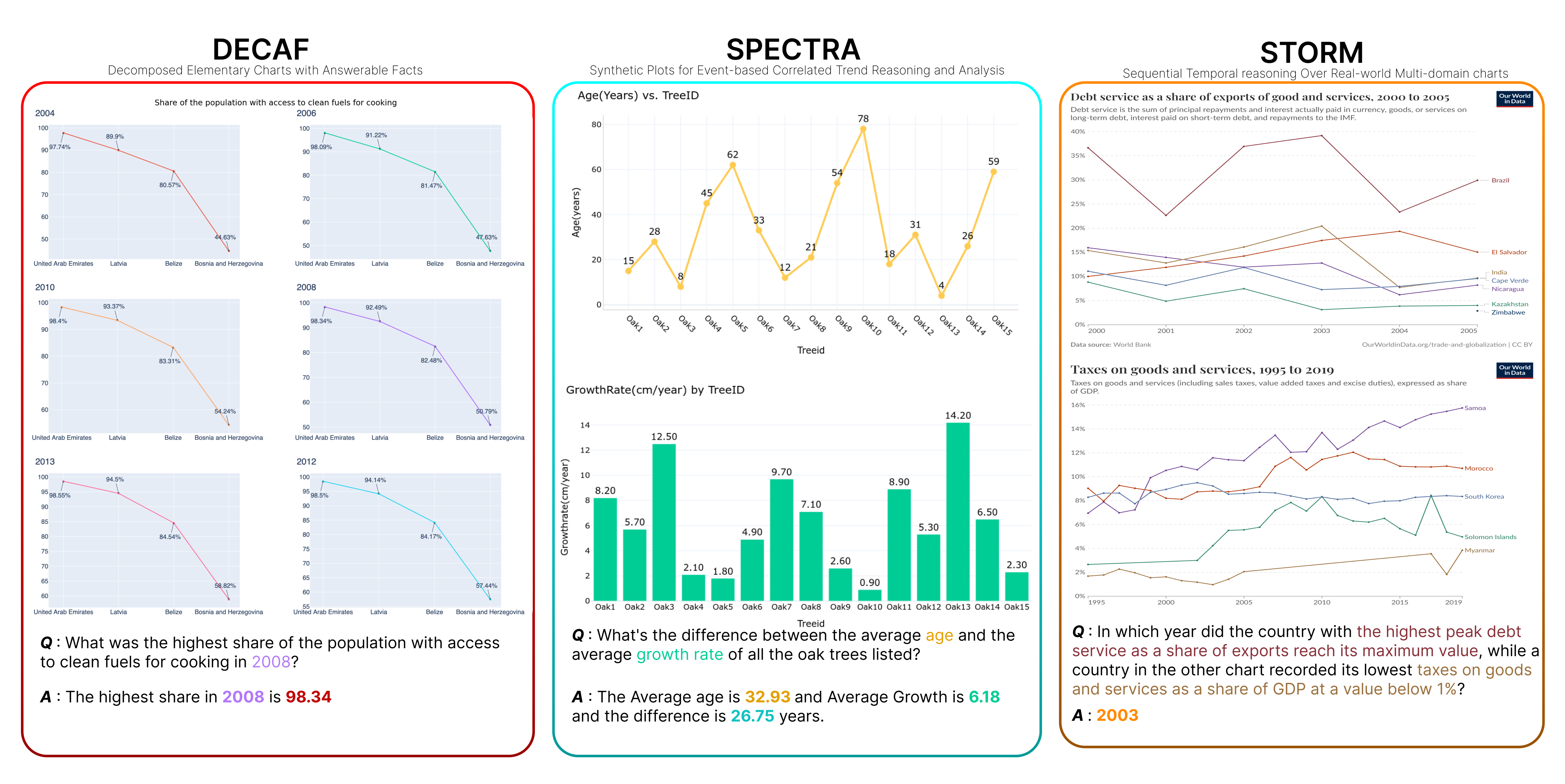}
\caption{Illustrative examples from our \InterChart{} benchmark: DECAF, SPECTRA, and STORM. The DECAF example shows a decomposed version of a chart similar to one found in STORM.}
\label{fig:enter-label}
\end{figure*}



\section{The \InterChart{} Benchmark}

We introduce \InterChart{} to systematically evaluate how reasoning difficulty, chart diversity, and visual complexity affect performance in vision-language models (VLMs). The benchmark contains 5,214 validated question-answer (QA) pairs divided into three subsets: \emph{DECAF}, \emph{SPECTRA}, and \emph{STORM}. These subsets represent distinct levels of real-world chart interpretation difficulty. ~\ref{app:flowcharts} summarizes the benchmark construction and annotation workflow for all three subsets, with detailed pipeline diagrams in Figures~\ref{fig:Flowchart_for_S1}, \ref{fig:s2}, and \ref{fig:s3}, and corresponding generation algorithms in ~\ref{app:data_gene_algo}.


\subsection{DECAF - Decomposed Elementary Charts with Answerable Facts}

The \emph{DECAF} subset establishes a foundation for evaluating baseline chart understanding. It includes both real and synthetic charts that represent single variables with minimal visual clutter. The QA tasks focus on factual lookup, comparisons, and parallel reasoning across clearly presented data.

\begin{table}[h]
\begin{small}
\centering
\renewcommand{\arraystretch}{1.2}
\setlength{\tabcolsep}{5pt}
\begin{tabular}{l r|l r}
\toprule
\multicolumn{4}{l}{\textbf{\textit{DECAF Distributions}}} \\
\midrule
\multicolumn{2}{l}{\textbf{Chart Type}} & \multicolumn{2}{l}{\textbf{Original Chart Sources}} \\
\midrule
\textit{Line}            & 22  & \textit{ChartQA}     & 153 \\
\textit{Horizontal Bar}  & 52  & \textit{DVQA}        & 70  \\
\textit{Vertical Bar}    & 149 & \textit{ChartInfo}   & 27  \\
\textit{Box Plot}        & 58  & \textit{ChartLlama}  & 105 \\
\textit{Heat Map}        & 37  &                      &     \\
\textit{Dot}             & 37  &                      &     \\
\midrule
\multicolumn{2}{l}{\textbf{\shortstack{\\QA Generation Methods}}} & \multicolumn{2}{l}{\textbf{Total}} \\
\midrule
\textit{Original QA}     & 665  & \textbf{QA Pairs}         & \textbf{2,809} \\
\textit{Table-LLM}       & 1,467 & \textbf{Original Charts}  & \textbf{355} \\
\textit{Table-SQL-LLM}   & 677  & \textbf{Decomposed Charts} & \textbf{1,188} \\
\bottomrule
\end{tabular}
\caption{Summary of chart types, sources, QA generation, and totals for \textit{DECAF}.}
\label{tab:s1_summary}
\end{small}
\end{table}

\paragraph{Chart Construction} We selected compound charts from ChartQA~\cite{masry2022chartqa}, ChartLlama~\cite{han2023chartllama}, ChartInfo~\cite{davila2025chartinfo}, and DVQA~\cite{kafle2018dvqa}, ensuring diverse sources of real-world chart styles and semantics. These charts span common types such as vertical and horizontal bar plots, line charts, box plots, dot plots, and heatmaps, covering a wide spectrum of visual encodings frequently used in analytical documents. To support reasoning at a granular level, we aimed to isolate atomic facts from multi-variable visuals. When necessary, we used DePlot~\cite{liu2023deplot} to regenerate missing tables from raw chart images, ensuring data fidelity and completeness. We then employed a custom decomposition script that extracted individual rows from these tables, aligned them with chart legends and axis labels, and rendered simplified single-variable charts using Plotly. This transformation allowed us to break down dense compound visuals into interpretable units, promoting focused reasoning over elementary visual elements. The complete data decomposition pipeline is illustrated in Appendix - Figure~\ref{fig:Flowchart_for_S1} and Algorithm~\ref{alg:sql-sampling}. This resulted in 355 compound charts and 1,188 decomposed charts.

\paragraph{QA Generation} We employed a SQL-based sampling strategy to generate table slices. We then used deterministic query templates and Gemini 1.5 pro to create natural language QA pairs, including both chart- and table-derived prompts. A filtering process reduced over 36,000 pairs to 5,800 candidates, followed by manual review to finalize 2,809 QA pairs. Table~\ref{tab:s1_summary} details the chart types, sources, and QA generation methods in \emph{DECAF}.


\subsection{SPECTRA - Synthetic Plots for Event-based Correlated Trend Reasoning and Analysis}

The \emph{SPECTRA} subset evaluates a model's ability to integrate distributed information across visually distinct but thematically aligned synthetic charts. These scenarios simulate real-world reasoning, such as interpreting relationships between variables that evolve over time or across regions.

\paragraph{Chart Construction} We created structured tables with shared axes to emulate real-world analyses (e.g., linking urban green space with happiness), ensuring that each table reflected plausible entity relationships across dimensions such as time, geography, or category. These base tables served as input to a two-step synthetic chart construction pipeline. First, we used Gemini 1.5 Pro to generate tabular data with natural variability across rows and columns, guided by template-based prompt scaffolds that preserved semantic consistency while allowing domain shifts (e.g., GDP vs. life expectancy). Second, the structured tables were rendered into visually diverse charts using a human-in-the-loop chart generation module. This included manual oversight to ensure balanced axis scales, legend consistency, and type diversity (e.g., bar-line overlays, multi-axis scales). The resulting charts preserved shared axes across pairs, promoting alignment in subsequent QA tasks. The corresponding generation flow is detailed in Appendix - Figure~\ref{fig:s2} and Algorithm~\ref{alg:synthetic-simulation}. Through this pipeline, we generated synthetic yet realistic chart combinations that encouraged event-based correlation and cross-variable reasoning.
\vspace{-2pt}
\paragraph{QA Generation} We prompted the model to generate questions targeting \textit{low-level reasoning}, such as computing totals or averages; \textit{trend analysis}, including directional inferences and value predictions; and \textit{scenario-based inference}, such as multi-condition comparisons. We used a Python-enabled LLM agent to validate answers through intermediate computation before converting outputs into natural language. After validation, the \emph{SPECTRA} subset contains 1,717 QA pairs across 333 visual context sets and 870 unique charts. Table~\ref{tab:s2s3_summary} provides detailed distributions.

\begin{figure*}[t]
\centering
\includegraphics[width=0.9\linewidth]{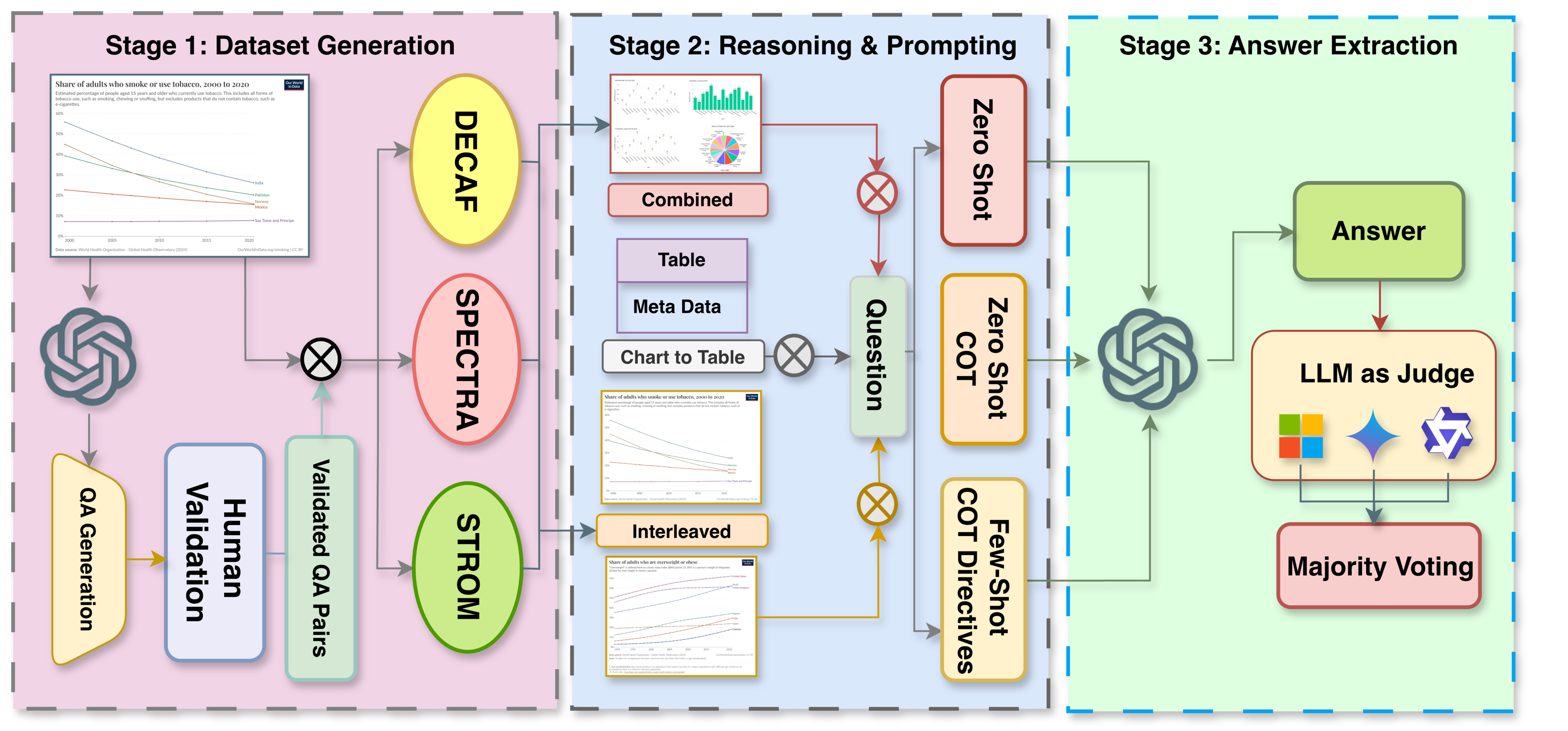}
\caption{Overview of the \textbf{\InterChart{}} Benchmark Pipeline.}
\label{fig:overview}
\end{figure*}

\subsection{STORM - Sequential Temporal reasoning Over Real-world Multi-domain charts}

The \emph{STORM} subset probes the upper limits of current VLM capabilities. It contains complex real-world line chart pairs with diverse styles and domains. These chart combinations reflect realistic analysis settings such as economic reports, environmental trends, and public health dashboards.

\begin{table}[h]
\begin{small}
\centering
\renewcommand{\arraystretch}{1.2}
\setlength{\tabcolsep}{6pt}
\begin{tabular}{l r | l r}
\toprule
\multicolumn{4}{l}{\textbf{\textit {SPECTRA \& STORM Distribution}}} \\
\midrule
\multicolumn{2}{l}{\textbf{SPECTRA}} & \multicolumn{2}{l}{\textbf{STORM}} \\
\midrule
\textit{Correlated}  & 1,481 & \textit{Range Estimation}         & 198 \\
\textit{Independent} & 245   & \textit{Abstract Numerical} & 275 \\
                     &       & \textit{Entity Inference}          & 295 \\
\midrule
\multicolumn{4}{l}{\textbf{Totals}} \\
\midrule
\textbf{QA Pairs}         & \textbf{1,717} & \textbf{QA Pairs}          & \textbf{768} \\
\textbf{Context Sets}     & \textbf{333}   & \textbf{Original Charts}   & \textbf{324} \\
\textbf{Unique Charts}    & \textbf{870}   & \textbf{Unique Images}     & \textbf{648} \\
\bottomrule
\end{tabular}
\caption{Distribution of question types and overall counts in \textit{SPECTRA} and \textit{STORM}.}
\label{tab:s2s3_summary}
\end{small}
\end{table}

\newcommand{\OWID}{Our World in Data\footnote{* Our World in Data: \url{https://ourworldindata.org/}}}

\paragraph{Chart Collection} We crawled charts and associated metadata from the \OWID{*}  repository. Using semantic cues and metadata attributes, we applied a semantic pairing module to group charts into coherent visual contexts that share related entities across time. The pairing process identified candidate chart pairs with aligned topics or axes, such as GDP and healthcare spending over the same time period. Each candidate pair was manually reviewed to ensure contextual relevance and analytical coherence. The chart construction pipeline followed the \textit{STORM} algorithmic design outlined in Appendix - Algorithm~\ref{alg:realm}, incorporating structured metadata extraction, entity alignment, and refinement steps to yield 324 validated chart sets comprising 648 distinct images. A visual overview of this pipeline is provided in Appendix - Figure~\ref{fig:s3}.

\paragraph{QA Curation} We used Gemini 2.5 Pro to generate candidate QA pairs grounded in both the chart images and their metadata, while Gemini~1.5~Pro was consistently used across all subsets (\textit{DECAF}, \textit{SPECTRA}, and \textit{STORM}) for model evaluation to maintain benchmarking uniformity. The QA generation process focused on multi-step reasoning that spans both charts in a pair, including contextual range estimation, numerical comparisons, temporal trend evaluation, and entity-based inference. Human annotators refined the generated QA pairs to ensure clarity, correctness, and depth of reasoning. Each pair was reviewed, categorized, and finalized through a collaborative validation loop, as described in Algorithm~\ref{alg:realm}. The resulting \emph{STORM} subset includes 768 QA pairs across the verified chart sets. Table~\ref{tab:s2s3_summary} summarizes the distribution of question types and chart contexts. 

\paragraph{Chart Type Rationale} 
We focused the \textit{STORM} subset on line charts because they dominate real-world analytical settings involving temporal reasoning. Domains such as public health, macroeconomics, and environmental science often present related time series (e.g., GDP vs.\ CO$_2$ emissions) using side-by-side line charts. By restricting to this chart type, we ensured consistent axis alignment and minimized confounding factors from mixed visual styles, allowing us to construct multi-step aggregation and temporal inference questions while preserving semantic interpretability.


\subsection{\InterChart{} Verification}

We implemented a multi-stage verification pipeline that combined automated filtering and human validation to ensure the quality of \InterChart{}.

We first used LLM-based acceptability checks to remove ambiguous or malformed QA pairs. Next, a team of 6 graduate-level annotators manually reviewed each item in DECAF and SPECTRA, ensuring correctness and diversity. Two graduate-level annotators independently verified every QA pair of STORM, with arbitration used to resolve disagreements.

\begin{table}[h]
\begin{small}
\centering
\renewcommand{\arraystretch}{1.2}
\setlength{\tabcolsep}{3pt}
\begin{tabular}{lccc}
\toprule
\textbf{} & \textbf{QA Samples} & \textbf{DECAF} & \textbf{SPECTRA} \\
\midrule
Pre  & 13,000 & 5,800 & 4,800 \\
Post & 5,214  & 2,809 & 1,717 \\
\% Drop & 59.9\% & 51.6\% & 64.2\% \\
\bottomrule
\end{tabular}
\caption{\InterChart{} human filtering statistics showing QA sample counts before and after manual verification for subsets \textit{DECAF} and \textit{SPECTRA}.}
\label{tab:filtering_stats}
\end{small}
\end{table}

Table~\ref{tab:filtering_stats} shows filtering statistics for the \emph{DECAF} and \emph{SPECTRA} subsets, revealing retention rates after manual curation. Table~\ref{tab:kappa_stats} shows the inter-annotator agreement for the \emph{STORM} subset, measured using Cohen’ Kappa. We achieved a agreement score of 70.63\%, reflecting consistent annotations for complex multi-chart reasoning.


\begin{table}[h]
\begin{small}
\centering
\setlength{\tabcolsep}{3pt}
\begin{tabular}{lccc}
\toprule
\textbf{}     & \textbf{QA Samples} & \textbf{Cohen’s $\kappa$} & \textbf{Jaccard Index} \\
\midrule
\textbf{Overall} & 768 & 70.63\% & 94.75\%                \\
\bottomrule
\end{tabular}
\caption{Overall inter‐annotator agreement (Cohen’s $\kappa$) for the STORM  annotated subsets.}
\label{tab:kappa_stats}
\end{small}
\end{table}

\textbf{Final Dataset Overview:} \InterChart{} includes \textbf{5,214 validated QA pairs} across \textbf{1,012 multi-chart contexts} and \textbf{2,706 unique chart images}. These examples span diverse reasoning types, visual structures, and real-world complexities, making \InterChart{} a comprehensive diagnostic resource for evaluating multi-chart visual question answering.

\section{Experiments}
\label{sec:modeling}

We benchmark visual reasoning on \InterChart{} using a diverse set of vision-language models (VLMs) and multiple input strategies. Our experiments address four core questions: (1) Does chart decomposition improve accuracy? (2) How does visual complexity affect multi-chart reasoning? (3) Can prompt engineering enhance performance? (4) Do structured tables offer an advantage over direct visual inputs?

\paragraph{VLMs} We evaluate both closed- and open-source VLMs. \textbf{Closed-source models} include Google Gemini 1.5 Pro~\cite{Gemini15} and OpenAI GPT-4o Mini~\cite{GPT4oMini}. \textbf{Open-source models} include Qwen2-VL-7B-Instruct~\cite{yang2024qwen2}, MiniCPM-V-2\_6~\cite{hu2024minicpm}, InternVL-2-8B~\cite{chen2024internvl}, and Idefics3-8B-LLaMA3~\cite{laurenccon2024building}. We also include DePlot~\cite{liu2023deplot} and Chart-to-Text~\cite{kantharaj2022chart} to assess reasoning over structured outputs.

\subsection{Evaluation Pipelines}

We compare two reasoning pathways: direct chart-based VQA and a chart-to-table pipeline using intermediate structured representations.

\begin{table*}[t]
\small
\centering
\renewcommand{\arraystretch}{1}
\setlength{\tabcolsep}{2.75pt}
\begin{tabular}{lcccccccccccc}
\toprule
\textbf{Model} & \multicolumn{4}{c}{\textbf{Zero-Shot}} & \multicolumn{4}{c}{\textbf{Zero-Shot CoT}} & \multicolumn{4}{c}{\textbf{Few-Shot CoT$_D$}} \\
& \textbf{\footnotesize Net} & \textbf{\scriptsize DECAF} & \textbf{\scriptsize SPECTRA} & \textbf{\scriptsize STORM} & \textbf{\footnotesize Net} & \textbf{\scriptsize DECAF} & \textbf{\scriptsize SPECTRA} & \textbf{\scriptsize STORM} & \textbf{\footnotesize Net} & \textbf{\scriptsize DECAF} & \textbf{\scriptsize SPECTRA} & \textbf{\scriptsize STORM} \\
\midrule
\multicolumn{13}{c}{\textit{Combined Visual Context Image}} \\
\midrule
GPT-4o-mini         & 44.8 & 59.3 & 45.6 & 29.7 & 48.5 & 68.3 & 47.9 & 29.4 & 48.8 & 68.6 & 47.2 & 30.6 \\
Gemini-1.5-Pro      & \textbf{53.0} & \textbf{65.2} & \textbf{59.1} & 34.8 & \textbf{55.0} & \textbf{71.6} & \textbf{58.5} & 34.9 & \textbf{56.3} & \textbf{73.9} & \textbf{61.5} & \textbf{33.7} \\
Qwen2-VL-7B         & 37.3 & 50.2 & 32.8 & 28.9 & 41.8 & 59.9 & 37.3 & 28.4 & 40.4 & 56.3 & 37.0 & 27.9 \\
MiniCPM-V-2\_6      & 34.3 & 52.2 & 32.4 & 21.5 & 35.3 & 52.7 & 31.9 & 21.3 & 32.4 & 48.7 & 30.1 & 18.6 \\
InternVL-2-8B       & 30.4 & 40.0 & 26.6 & 24.8 & 32.3 & 45.2 & 28.2 & 23.6 & 31.6 & 46.3 & 27.3 & 21.2 \\
Idefics3-8B-Llama3  & 23.2 & 39.3 & 19.4 & 11.1 & 23.8 & 38.8 & 19.6 & 13.1 & 25.9 & 35.7 & 25.1 & 17.1 \\
\midrule
\textbf{Mean} & 37.2 & 51.0 & 36.0 & 25.1 & 39.5 & 56.1 & 37.2 & 25.1 & 39.2 & 55.0 & 38.0 & 24.9\\
\midrule
\multicolumn{13}{c}{\textit{Interleaved Visual Context}} \\
\midrule
GPT-4o-mini         & 41.9 & 44.4 & 50.0 & 31.5 & 44.5 & 51.5 & 50.3 & 31.9 & 44.4 & 51.7 & 50.4 & 31.1 \\
Gemini-1.5-Pro      & 52.7 & 64.7 & 57.4 & \textbf{36.0} & 54.1 & 68.1 & 57.8 & \textbf{36.4} & 54.2 & 70.3 & 59.6 & 32.9 \\
Qwen2-VL-7B         & 37.0 & 49.3 & 32.9 & 28.9 & 39.4 & 52.8 & 38.7 & 26.7 & 36.1 & 47.9 & 35.2 & 25.2 \\
MiniCPM-V-2\_6      & 37.1 & 49.3 & 36.8 & 25.2 & 36.6 & 49.6 & 36.2 & 24.2 & 35.5 & 48.1 & 35.1 & 23.5 \\
\midrule
\textbf{Mean} & 42.2 & 51.9 & 44.3 & 30.4 & 43.7 & 55.5 & 45.8 & 29.8 & 42.6 & 54.5 & 45.1 & 28.2\\
\bottomrule
\end{tabular}
\caption{Accuracies using our evaluation method with majority voting of evaluators on all models and prompting strategies. Results are grouped by visual context format (top: Combined, bottom: Interleaved), and broken down by set type (DECAF, SPECTRA, STORM) and strategy (Zero-Shot, Zero-Shot CoT, Few-Shot CoT with Directives). Net scores refer to the mean score of the model across different subsets.}
\label{tab:model_evaluation_extended}
\end{table*}

\paragraph{Direct Chart Question Answering}
We test two visual formats: (i) \textbf{Combined}, where charts are stitched into a unified image, and (ii) \textbf{Interleaved}, where charts are passed sequentially. For DECAF, we also evaluate original compound charts to quantify gains from simplification.

Prompting styles include \textbf{Zero-Shot}, \textbf{Zero-Shot CoT} (stepwise reasoning), and \textbf{Few-Shot with Directives}~\cite{flowchartqa}, which gives structured step-level guidance. Due to input size limits, InternVL and Idefics3 are excluded from interleaved inputs.

\paragraph{Table as Intermediate Representation}
This setup evaluates whether structured conversion aids reasoning. It includes: (1) \textit{Chart-to-Table Conversion}, where models extract metadata and tables from images, and (2) \textit{Table-Based QA}, where models answer using these tables via CoT prompts. We compare Gemini 1.5 Pro, Qwen2-VL, and MiniCPM. To address DePlot’s title extraction issues, we augment it using Gemini title generation, yielding an improved hybrid we term \textbf{DePlot++}. This isolates the benefit of structure vs. visual inputs under matched prompts.

\paragraph{Evaluation Strategy}
\label{sec:evaluation}
We use LLM-based semantic judges to score answers beyond exact string matching, supporting paraphrases, numerics, and unit variations if reasoning is correct. Evaluators include \textbf{Gemini 1.5 Flash (8B)}~\cite{Gemini15}, \textbf{Phi 4}~\cite{phi4}, and \textbf{Qwen2.5-7B-Instruct}~\cite{yang2024qwen2technicalreport}. These models were selected to ensure architectural diversity across families (Google Gemini, Microsoft Phi, and Alibaba Qwen), balanced parameter scales between 7B-8B for efficiency and semantic depth, and empirical reliability validated through agreement testing. Each receives the question, reference answer, and model output, and returns a binary correctness score along with its reasoning. Final scores use majority voting. A broader discussion comparing this evaluation framework with automatic text-based metrics such as BLEURT, MoverScore, and QuestEval is provided in~\ref{appendix:automatic_metrics}.


To validate the majority voting agreement, we benchmarked 10,000 sampled responses. In over \textbf{78.67\%} of cases, all three evaluators agreed on a common answer. Per-model breakdowns appear in~\ref{app:appendix_d}.


\section{Results and Analysis}
\label{sec:results}

We analyze performance on \InterChart{} across visual input formats, prompting strategies, and subset difficulty levels by answering targeted questions that highlight emerging trends, model strengths, and failure modes. Tables~\ref{tab:model_evaluation_extended} through~\ref{tab:s3_decomposition} summarize these results.

\subsection{Performance across Chart Subsets}

\paragraph{Do Interleaved Charts Help Models Perform Better than Combined Charts?}
Not consistently. As shown in Table~\ref{tab:model_evaluation_extended}, interleaving charts sometimes improves performance but often leads to minimal or negative changes. For example, Gemini-1.5 Pro improves slightly in STORM from 34.8\% to 36.0\% but drops from 65.2\% to 64.7\% in DECAF. Qwen2-VL decreases in DECAF (50.2\% to 49.3\%) and SPECTRA (32.8\% to 32.9\%). MiniCPM improves modestly in STORM (21.5\% to 25.2\%). These results suggest interleaving may help with visual clutter in complex charts but does not offer consistent benefits across all subsets.

\paragraph{Does Decomposing Charts Improve Model Accuracy?}
Yes. As shown in Table~\ref{tab:chart_to_table_accuracy}, converting charts into structured tables improves accuracy in many cases. Gemini-1.5 Pro achieves 69.9\% accuracy using structured DECAF tables, outperforming both DePlot (54.3\%) and C2T (43.8\%). DePlot++ further improves performance to 63.2\% by enhancing title and metadata alignment. Qwen2-VL and MiniCPM also benefit modestly, though their scores remain lower (50.1\% and 33.8\%, respectively). These results suggest that SQL-based decomposition paired with table-driven reasoning can improve clarity and support more accurate inference compared to image-only inputs.

\paragraph{Why Do Models Perform Poorly on Real-World Multi-Chart Tasks?}
As seen in Table~\ref{tab:model_evaluation_extended}, accuracy drops sharply in the STORM subset. Gemini-1.5 Pro falls to 34.8\%, Qwen2-VL to 28.9\%, and MiniCPM-V-2\_6 to 21.5\%. These real-world chart pairs demand semantic alignment and temporal synthesis. Table~\ref{tab:s3_decomposition} shows abstract numerical reasoning is hardest (15.6\%), followed by range estimation (33.4\%) and entity inference (39.1\%). These declines reflect the challenge of integrating misaligned metadata, irregular axes, and domain-specific trends across diverse visual styles.

\paragraph{Do Models Generalize Well from Synthetic to Real-World Chart Distributions?}
No. Table~\ref{tab:model_evaluation_extended} shows a consistent drop in performance from SPECTRA to STORM across all models. Gemini-1.5 Pro declines from 59.1\% in SPECTRA to 34.8\% in STORM. Qwen2-VL drops from 32.8\% to 28.9\%, and MiniCPM-V-2\_6 from 32.4\% to 21.5\%. These results suggest that while models handle synthetic trend-based reasoning to some extent, they struggle to transfer those skills to real-world chart pairs that involve domain shifts, visual diversity, and temporal reasoning.

\subsection{Effect of VLMs}

\paragraph{Why Does Gemini-1.5 Pro leads within the tested baseline suite?}
Gemini-1.5 Pro consistently leads across all subsets and prompting strategies. As shown in Table~\ref{tab:model_evaluation_extended}, it scores 65.2\% in DECAF, 59.1\% in SPECTRA, and 34.8\% in STORM-well ahead of all other models. GPT-4o-mini is the next best, but lags in STORM (29.7\%). Open-source models like Qwen2 and MiniCPM perform reasonably in DECAF but decline sharply on harder subsets. Gemini’s strength likely stems from its training on structured inputs and strong instruction-following capabilities. GPT-4o achieved performance levels that closely approach those of Gemini-1.5 Pro, particularly in the \textit{STORM} subset that emphasizes semantic abstraction and temporal reasoning (see~\ref{tab:gpt4o_appendix}).

\paragraph{How Do Open-Source Models Compare Across Subsets?}
Open-source models perform well in DECAF but struggle in SPECTRA and STORM. Qwen2-VL-7B drops from 50.2\% in DECAF to 32.8\% in SPECTRA and 28.9\% in STORM. MiniCPM-V-2\_6 shows a similar decline: 52.2\% $\rightarrow$ 32.4\% $\rightarrow$ 21.5\%. InternVL and Idefics3 perform lower across all subsets, particularly in STORM. These trends point to challenges in generalization, especially when models face domain shifts and complex temporal reasoning.

\subsection{Effect of Strategies}

\paragraph{Which Prompting Strategies Work Best Across Subsets?}

\begin{table}[t]
\small
\centering
\setlength{\tabcolsep}{4pt}
\begin{tabular}{lcccc}
\toprule
\textbf{Model} & \textbf{\scriptsize DECAF} & \textbf{\scriptsize SPECTRA} & \textbf{\scriptsize STORM} & 
{\textbf{\scriptsize DECAF$_o$}}
 \\
\midrule
C2T              & 43.8 & 46.3 & 14.7 & 62.6 \\
Gemini-1.5-Pro   & \textbf{69.9} & \textbf{68.1} & \textbf{29.5} & \textbf{76.0} \\
Deplot           & 54.3 & 57.9 & 22.2 & 63.8 \\
Deplot++         & 63.2 & 58.1 & 23.6 & 61.9 \\
MiniCPM-V-2\_6   & 33.8 & 22.1 & 12.2 & 35.6 \\
Qwen2-VL-7B      & 50.1 & 34.3 & 18.4 & 52.4 \\
\bottomrule
\end{tabular}
\caption{Accuracies from the chart-to-table prompting and rendering strategies for \textit{DECAF}, \textit{SPECTRA}, \textit{STORM}, and \textit{DECAF} compound charts: \textit{DECAF$_o$}.}
\label{tab:chart_to_table_accuracy}
\end{table}

Few-Shot Chain-of-Thought with Directives generally yields the highest accuracy across models and subsets. Table~\ref{tab:model_evaluation_extended} shows Gemini-1.5 Pro improves from 65.2\% (Zero-Shot) to 71.6\% (Zero-Shot CoT), and further to 73.9\% using Few-Shot CoT$_D$ in DECAF. Qwen2-VL follows a similar pattern, improving from 50.2\% to 59.9\%, before dropping slightly to 56.3\%. While MiniCPM sees minor gains with CoT, it drops slightly under Few-Shot CoT$_D$. Overall, structured prompting helps most in DECAF and SPECTRA, but offers limited advantage in STORM due to its high complexity.

\paragraph{Does Chain-of-Thought (CoT) Consistently Help?}
Mostly in simpler subsets. Table~\ref{tab:model_evaluation_extended} shows that CoT improves performance in DECAF and SPECTRA but offers limited benefit in STORM. For example, Gemini-1.5 Pro jumps from 65.2\% to 71.6\% in DECAF and from 59.1\% to 58.5\% in SPECTRA. Qwen2-VL improves from 50.2\% to 59.9\% in DECAF, and MiniCPM sees only a marginal gain (52.2\% to 52.7\%). In STORM, scores remain largely unchanged or even decline slightly, indicating that verbal reasoning alone cannot compensate for high visual and semantic complexity.

\subsection{Effect of Intermediate Representation}


\paragraph{How Do Different Table Extraction Methods Compare?}
DePlot++ consistently outperforms DePlot in DECAF and SPECTRA. As shown in Table~\ref{tab:chart_to_table_accuracy}, DePlot++ achieves 63.2\% in DECAF and 58.1\% in SPECTRA, compared to 54.3\% and 57.9\% with DePlot. 

\begin{table}[h]
\small
\centering
\begin{tabular}{lcc}
\toprule
\textit{\emph{DECAF} Chart Type} & \textbf{Mean} & \textbf{Best} \\
\midrule
\textbf{DECAF-Decomposition} & & \\
Line            & 39.66 & 57.76 \\
Horizontal Bar  & 50.95 & 73.36 \\
Vertical Bar    & 56.17 & 78.63 \\
Box Plot        & \textbf{64.3}  & \textbf{84.23} \\
Heat Map        & 55.36 & 81.35 \\
Dot             & 58.24 & 78.63 \\
\bottomrule
\end{tabular}
\caption{Distribution of Accuracies for Chart Decomposition Approach for \textit{DECAF}.}
\label{tab:s1_decomposition}
\end{table}

\begin{table}[h]
\small
\centering
\begin{tabular}{lcc}
\toprule
\textit{\emph{SPECTRA} Question Category} & \textbf{Mean} & \textbf{Best} \\
\midrule
\textbf{DECAF-Decomposition} & & \\
Correlated   & 39.49 & 67.43 \\
Independent  & \textbf{43.22} & \textbf{73.47} \\
\bottomrule
\end{tabular}
\caption{Distribution of Accuracies for Question Categorization Approach for \textit{SPECTRA}.}
\label{tab:s2_decomposition}
\end{table}

\noindent This improvement reflects better title and axis alignment, which helps structured models parse tabular input more accurately. The gains are modest but consistent, affirming the importance of clean preprocessing and metadata fidelity.

\paragraph{When Do Structured Tables Hurt Performance Instead of Helping?}
In STORM. As shown in Tables~\ref{tab:chart_to_table_accuracy} and~\ref{tab:model_evaluation_extended}, structured representations often degrade accuracy on complex real-world charts. Gemini-1.5 Pro drops from 34.8\% with visual inputs to 29.5\% using tables. C2T performs even worse at 14.7\%. These trends suggest that tables cannot capture semantic and temporal alignment across axes, which are critical for accurate reasoning in real-world multi-chart settings.

\subsection{Effect of Chart Types, Question Category, and Reasoning Type}

\paragraph{Which Chart Types Are Easier or Harder in DECAF?}
According to Table~\ref{tab:s1_decomposition}, box plots (64.3\%) and dot plots (58.24\%) are the easiest for models to interpret, followed by vertical bars (56.17\%). Line charts (39.66\%) and horizontal bars (50.95\%) yield lower accuracy, likely due to visual ambiguity in axis orientation and overlapping labels. These results suggest that models perform best when the chart layout is clean and the data encoding is visually distinct.

\paragraph{Which Question Types Are Easier in SPECTRA?}

Table~\ref{tab:s2_decomposition} shows that independent questions achieve higher accuracy (43.22\%) than correlated ones (39.49\%). 
\begin{table}[h]
\small
\centering
\setlength{\tabcolsep}{6pt}
\begin{tabular}{lcccc}
\toprule
\textbf{\textit{STORM}} & \multicolumn{2}{c}{\textbf{Interleaved}} & \multicolumn{2}{c}{\textbf{Combined}} \\
\cmidrule{2-3} \cmidrule{4-5}
 \textit{Reasoning Type} & \textbf{Mean} & \textbf{Best} & \textbf{Mean} & \textbf{Best} \\
\midrule
Abstract Numerical & 13.6 & 23.7 & \textbf{15.6} & \textbf{25.5} \\
Entity Inference & \textbf{42.1} & \textbf{51.3} & 39.1 & 50.9 \\
Range Estimation & 31.2 & \textbf{52.3} & \textbf{33.4} & 47.5 \\
\bottomrule
\end{tabular}
\caption{Distribution of accuracies for reasoning type categorization in \textit{STORM}, comparing interleaved and combined visual formats.}
\label{tab:s3_decomposition}
\end{table}

\noindent This suggests that isolating variables in SPECTRA makes reasoning easier for models, while correlated questions introduce multi-step dependencies across charts that are harder to track and align.

\paragraph{How Do Reasoning Demands Shift from SPECTRA to STORM?}

Comparing Table~\ref{tab:s2_decomposition} and Table~\ref{tab:s3_decomposition} shows that models perform well on independent trend analysis in \textit{SPECTRA} but struggle with \textit{STORM}'s abstract and range-based questions. This decline reflects a shift from visual correlation to semantic and temporal abstraction, where simple alignment no longer suffices. Even models exceeding 70\% accuracy on SPECTRA’s independent questions drop below 35\% on STORM’s range estimation tasks, underscoring that \InterChart{} diagnoses distinct reasoning failures rather than cross-tier ranking.

\paragraph{How Consistent Are VLMs Across Chart Types?}
Model performance varies significantly across chart types. Table~\ref{tab:s1_decomposition} shows accuracies ranging from 39.66\% for line charts to 64.3\% for box plots. This variation suggests VLMs lack consistent chart-type generalization and are sensitive to layout complexity, axis orientation, and label density. Even high-performing models like Gemini show dips on dense or ambiguous formats, highlighting the need for chart-aware visual parsing.

\paragraph{How Do Reasoning Types Impact Performance in STORM?}
As shown in Table~\ref{tab:s3_decomposition}, reasoning type has a clear impact on accuracy in STORM. Entity inference yields the highest mean accuracy (42.1\% interleaved), followed by range estimation (33.4\%), and abstract numerical reasoning is lowest (13.6-15.6\%). Interleaved visual formats offer modest gains for entity and range tasks but have limited effect on abstract numerical reasoning, where semantic alignment and aggregation across charts remain key challenges.

\section{Comparison with Related Work}
Understanding visualizations through natural language has long been a goal in multimodal AI. Early chart-based VQA datasets such as FigureQA~\cite{kahou2017figureqa}, DVQA~\cite{kafle2018dvqa}, PlotQA~\cite{methani2020plotqa}, ChartQA~\cite{masry2022chartqa}, and ChartLlama~\cite{han2023chartllama} introduced benchmarks over synthetic or real-world plots, focusing on factual or reasoning questions in isolated visual contexts. Recent efforts like ChartInfo~\cite{davila2024chart} and SciGraphQA~\cite{li2023scigraphqa} extended this by incorporating structured data such as tables and graphs. However, these datasets center on single-chart scenarios and do not evaluate a model's reasoning ability across multiple, semantically related charts. Complementary work on multi-hop~\cite{deng-etal-2022-explicit} and graph-based QA~\cite{jin2024graph} has demonstrated that decomposing complex inputs into smaller units improves reasoning and interpretability. MultiChartQA~\cite{zhu-etal-2025-multichartqa} takes a step toward multi-chart reasoning through synthetic chart triplets and four structured tasks: direct, parallel, comparative, and sequential. While it offers controlled diagnostics, the benchmark uses uniformly styled charts with fixed layouts and semantics. It does not assess model performance under visual diversity, semantic drift, or layout complexity, which are standard features in real-world chart collections. Recent benchmarks such as InfoChartQA~\cite{lin2025infochartqa}, ChartMind~\cite{wei2025chartmind}, and ChartQAPro~\cite{masry2025chartqapro} have expanded chart understanding toward more realistic visual and linguistic settings. These datasets emphasize broader coverage and visual diversity but primarily address single-chart or loosely connected infographic reasoning. In contrast, \InterChart{} particularly its \emph{STORM} subset was explicitly designed to evaluate \emph{multi-chart} reasoning that demands semantic drift handling, temporal alignment, and multi-step integration across co-occurring charts. An illustrative STORM example involving temporally aligned chart pairs from \emph{Our World in Data} is provided in Appendix~\ref{fig:storm_example_pair}, demonstrating how models must correlate trends across independent metrics to infer temporally grounded answers.
\vspace{-2pt}
\par \textbf{\InterChart{}} addresses these gaps with a broader diagnostic lens. It introduces three subsets \emph{DECAF}, \emph{SPECTRA}, and \emph{STORM} spanning single-chart to real-world multi-chart reasoning under increasing difficulty and diversity. Unlike prior benchmarks, it combines synthetic and real-world charts to evaluate robustness to visual heterogeneity and abstraction. Additionally, it incorporates an LLM-based evaluation framework that assesses semantic correctness beyond string overlap. \InterChart{} thus serves both as a benchmark for evaluating performance and a diagnostic framework for identifying where current models fail in complex, multi-chart reasoning scenarios. To further clarify these distinctions, Appendix~\ref{appendix:comparison} presents a comparative table summarizing chart type coverage, reasoning scope, multi-chart design, semantic drift, temporal reasoning, and evaluation protocols across recent benchmarks (InfoChartQA, ChartMind, ChartQAPro, and \InterChart{}). This structured comparison highlights that \InterChart{} uniquely couples real-world multi-chart reasoning with semantic and temporal abstraction while maintaining diagnostic granularity through its LLM-based majority-voting evaluation.

\section{Conclusion and Future Directions}
We introduced \InterChart{}, a diagnostic benchmark for evaluating vision-language models (VLMs) on multi-chart visual reasoning. Structured across three progressively complex subsets \emph{DECAF}, \emph{SPECTRA}, and \emph{STORM}. \InterChart{} enables detailed analysis of model behavior under controlled visual transformations. Our findings show that while current VLMs perform well on simplified, decomposed visuals, their accuracy drops significantly when required to integrate or infer across visually complex, semantically misaligned chart sets. Rather than treating VQA as a binary success metric, \InterChart{} provides a controlled setting to explore \textit{why} models succeed or fail by varying presentation while holding semantic content constant. This enables diagnostic analysis of model robustness, attention mechanisms, and failure modes-offering insights relevant to model design, training strategies, and interface development.


In future work, we plan to expand \InterChart{} beyond traditional charts to include infographics, annotated scientific plots, and hybrid layouts. We also plan to extend the \textit{STORM} subset to heterogeneous chart combinations (e.g., line-bar or heatmap-scatter) to support broader reasoning analysis. We also aim to explore multilingual question sets and incorporate neuro-symbolic or retrieval-augmented approaches to support structured abstraction and cross-domain transfer. Furthermore, we plan to evaluate advanced prompting strategies such as self-consistency, reflection, and tree-of-thought (ToT) to assess their effectiveness in enhancing inter-chart reasoning. These directions can advance model transparency, scalability, and applicability in real-world decision-support settings.

\section*{Limitations}

\InterChart{} offers a flexible diagnostic framework but comes with limitations. First, our evaluations rely entirely on zero- and few-shot prompting due to resource constraints. This setup does not capture the full potential of models that might benefit from fine-tuning on chart-specific data. Second, all questions and visual content are English-only, which limits multilingual applicability. Additionally, the current version does not support spatial reasoning tasks such as bounding box grounding or region referencing. While we plan to add fine-grained annotations and structured parsing outputs in future versions, this study focuses solely on answer-level reasoning. Several potential extensions such as dynamic chart distillation, symbolic chart indexing, or JSON-based parsing supervision remain conceptual due to scope limitations. Despite these constraints, \InterChart{} lays a foundation for expanding multimodal evaluation toward structured, visual-first tasks. Future extensions could include layout-aware fine-tuning pipelines, grounded CoT prompting, and multimodal summarization agents tailored for multi-chart analytics.

\section*{Ethics Statement}
This work adheres to ethical standards in data collection, annotation, and reproducibility. All visual data used in \InterChart{} originate from publicly available or synthetically generated sources under permissible licenses. No sensitive or personally identifiable information is included. Annotations were conducted by graduate-level volunteers based in the United States and India, all of whom provided informed consent. To promote transparency and reproducibility, we will publicly release the full dataset, evaluation scripts, prompt templates, and annotation guidelines. All filtering heuristics and design decisions have been carefully documented to facilitate future research and benchmarking efforts. We also employed AI tools, including large language models, to assist with aspects of the project such as prompt development and explanatory text generation. All AI-generated outputs were reviewed and refined by human authors to ensure accuracy and clarity. Overall, this project reflects our commitment to data privacy, transparency, annotator welfare, and the responsible integration of AI tools throughout the research process.

\section*{Acknowledgments}
This research has been supported in part by the ONR Contract N00014-23-1-2364, and conducted as a collaborative effort between \textit{Arizona State University} and the \textit{University of Pennsylvania}. We gratefully acknowledge the Complex Data Analysis and Reasoning Lab at School of Augmented Intelligence, \textit{Arizona State University} for providing computational resources and institutional support. We also thank the anonymous reviewers for their thoughtful feedback and constructive suggestions. We extend special appreciation to our lab cat, Coco, whose presence helped both our team and our professor maintain just the right balance of focus and levity during deadlines.

We further acknowledge \textit{Varun Yerram, Prekshi Vyas, Mansi}, and \textit{Devanshi Garg} for their assistance during the early development phase of this project. Finally, we thank our parents for their unwavering encouragement and support throughout this project.

\bibstyle{acl_natbib}
\bibliography{custom} 
\clearpage

\appendix
\renewcommand\thesection{Appendix~\Alph{section}}

\section{Prompt Templates}
\label{app:prompts}
\subsection*{Zero-Shot Prompt}
\label{app:zero_shot}
\begin{tcolorbox}[title=Zero-Shot Prompt, colback=white, colframe=black, breakable]
Your task is to answer the question based on the given \{img\_word\}. Your final answer to the question should strictly be in the format \texttt{"Final Answer:" <final\_answer>}.

Question: \{question\}
\end{tcolorbox}

\subsection*{Zero-Shot Chain-of-Thought Prompt}
\label{app:zero_shot_cot}
\begin{tcolorbox}[title=Zero-Shot Chain-of-Thought Prompt, colback=white, colframe=black, breakable]
Your task is to answer the question based on the given \{img\_word\}. Your final answer to the question should strictly be in the format \texttt{"Final Answer:" <final\_answer>}. Let's work this out in a step by step way to be sure we have the right answer.

Question: \{question\}
\end{tcolorbox}

\subsection*{Data Extraction Prompt}
\label{app:data_extraction}
\begin{tcolorbox}[title=Data Extraction Prompt, colback=white, colframe=black, breakable]
Your task is to extract all data from the chart image provided. Make sure to include the chart's title. Output the data in a structured format. Ensure every data point is accurately captured and represented. Be meticulous and do not omit any information.

Think step by step. Identify the chart type to extract data accordingly.
\end{tcolorbox}

\subsection*{Table-Based Question Answering Prompt}
\label{app:table_qa}
\begin{tcolorbox}[title=Table-Based QA Prompt, colback=white, colframe=black, breakable]
You are tasked with answering a specific question. The answer must be derived solely from information provided, which is extracted from image(s) of chart(s). This information will include the data extracted from the chart, including the chart title. Your final answer to the question should strictly be in the format \texttt{"Final Answer:" <final\_answer>}. Let's work this out in a step-by-step way to be sure we have the right answer.

Data extracted from charts: \{tables\}

Question: \{question\}
\end{tcolorbox}

\subsection*{Chart Title Extraction Prompt}
\label{app:title_extraction}
\begin{tcolorbox}[title=Chart Title Extraction Prompt, colback=white, colframe=black, breakable]
Your task is to extract the main title of the chart image. The main title is typically located at the top of the chart, above the chart area itself, and describes the overall subject of the chart. The title usually describes what data is being presented, the time period, or the geographic location, if applicable.

If the chart does not have a discernible main title, your response should be \texttt{"Title: None"}. Otherwise, your response should be in the format \texttt{"Title: <title>"}.
\end{tcolorbox}

\subsection*{Few-Shot with Directives Prompt}
\label{app:few_shot}
\begin{tcolorbox}[title=Few-Shot with Directives Prompt, colback=white, colframe=black, breakable]
Your task is to answer a question based on a given \{img\_word\}. To ensure clarity and accuracy, you are required to break down the question into steps of extraction and reasoning. Your final answer should strictly rely on the visual information presented in the \{img\_word\}.

Here are a few directives that you can follow to reach your answer:

\textbf{Step 1: Identify Relevant Entities}  
First, identify the key entities or data points needed to answer the given question. These could be labels, categories, values, or trends in the chart or image.

\textbf{Step 2: Extract Relevant Values}  
Extract all necessary values related to the identified entities from the image. These values might be numerical (e.g., percentages, quantities) or categorical (e.g., labels, categories).

\textbf{Step 3: Reasoning and Calculation}  
Using the extracted values, apply logical reasoning and calculations to derive the correct answer. Explicitly state the reasoning process to ensure the steps leading to the final answer are understandable and correct.

\textbf{Step 4: Provide the Final Answer}  
Based on your reasoning, provide the final answer in the following format: \texttt{Final Answer: <final\_answer>}

Question: \{question\}
\end{tcolorbox}

\subsection*{LLM-as-a-Judge Prompt}
\label{app:llm_judge}
\begin{tcolorbox}[title=LLM-as-a-Judge Prompt, colback=white, colframe=black, breakable]
You will be given a question, the correct answer to that question (called the "Ground Truth answer"), and a student's attempt to answer the same question (called the "Student Written Answer"). Your task is to determine if the Student Written Answer is correct when compared to the Ground Truth answer.

\textbf{Instructions:}
\begin{itemize}
    \item The answer should be based solely on the provided information in the question and the Ground Truth answer.
    \item An answer is correct if it contains the same information as the Ground Truth answer, even if phrased differently.
    \item Ignore minor differences in wording or phrasing that do not change the meaning.
    \item If the Ground Truth answer is a number, consider the Student Written Answer correct if it is approximately equal (e.g., 20.24553 vs 20.24). State assumptions clearly.
    \item For range-based questions, accept answers within the correct range.
    \item Provide a short explanation inside \texttt{<reasoning>} tags.
    \item Output \texttt{<answer> 1 </answer>} if correct, or \texttt{<answer> 0 </answer>} if incorrect.
\end{itemize}

\textbf{Example:}  
Question: What is the color of water?  
Ground Truth answer: Pink  
Student Answer: \texttt{Final Answer: Water is colorless.}  

Response:  
\texttt{<reasoning> The student answer does not match the ground truth. </reasoning>}  
\texttt{<answer> 0 </answer>}  

Now, answer the following:  
Question: \{question\}  
Ground Truth answer: \{ground\_truth\}  
Student Written Answer: \{student\_answer\}
\end{tcolorbox}

\section{Flowcharts}
\label{app:flowcharts}

\begin{figure}[H]
    \centering
    \includegraphics[width=1\linewidth]{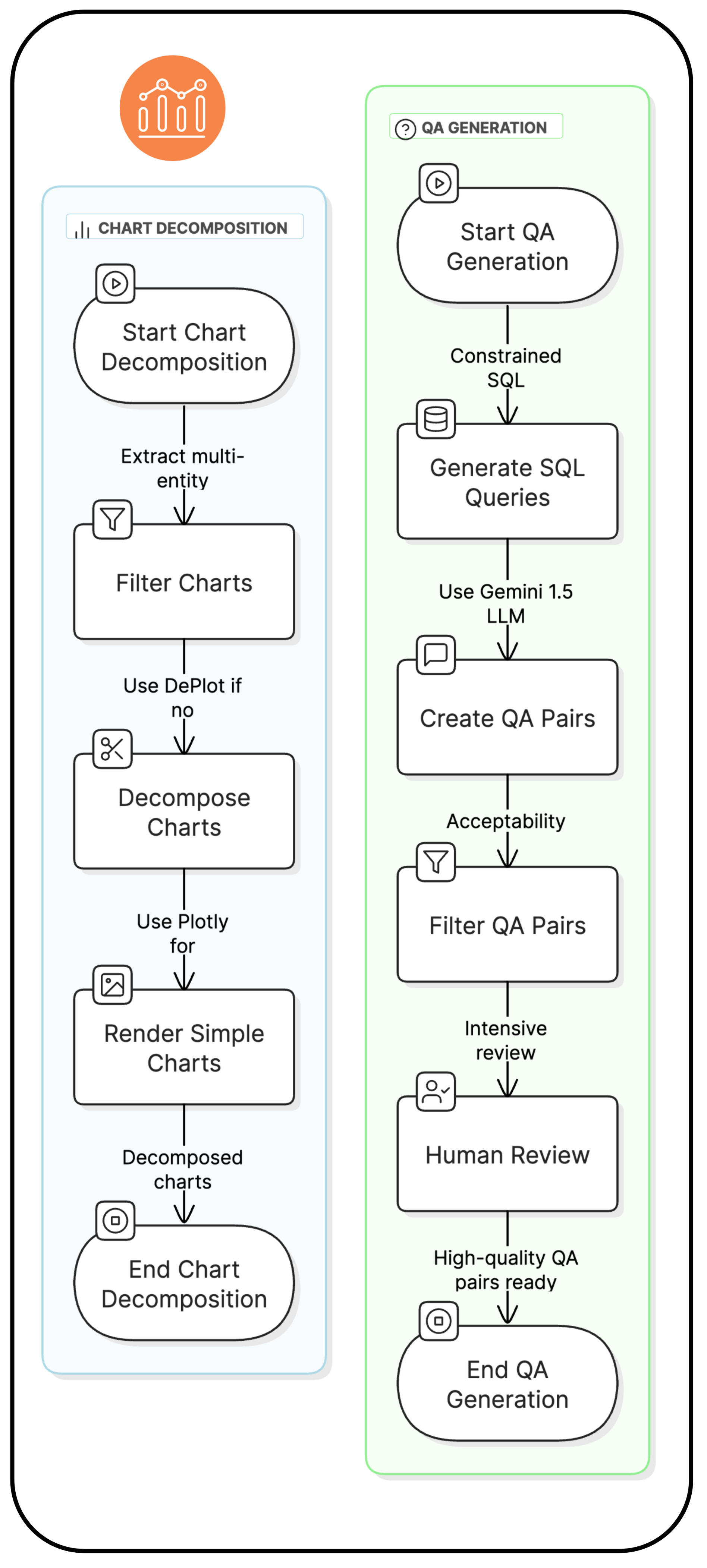}
    \caption{Pipeline for DECAF: Decomposing complex charts into simplified single-entity visuals and generating fact-based QA pairs.}
    \label{fig:Flowchart_for_S1}
\end{figure}

\begin{figure*}[htbp]
    \centering
    \begin{minipage}{0.48\textwidth}
        \centering
        \includegraphics[height=0.62\textheight]{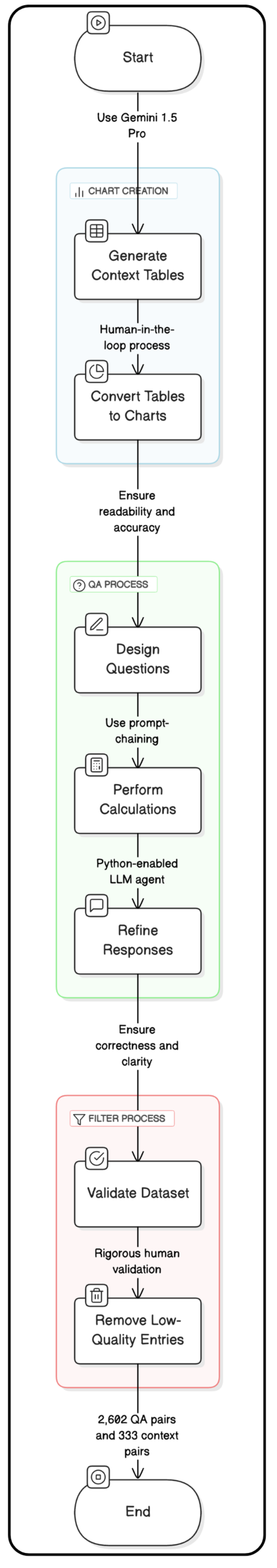}
        \caption{Pipeline for SPECTRA: Generating synthetic multi-chart contexts for correlated trend and scenario-based reasoning.}
        \label{fig:s2}
    \end{minipage}
    \hfill
    \begin{minipage}{0.48\textwidth}
        \centering
        \includegraphics[height=0.4\textheight]{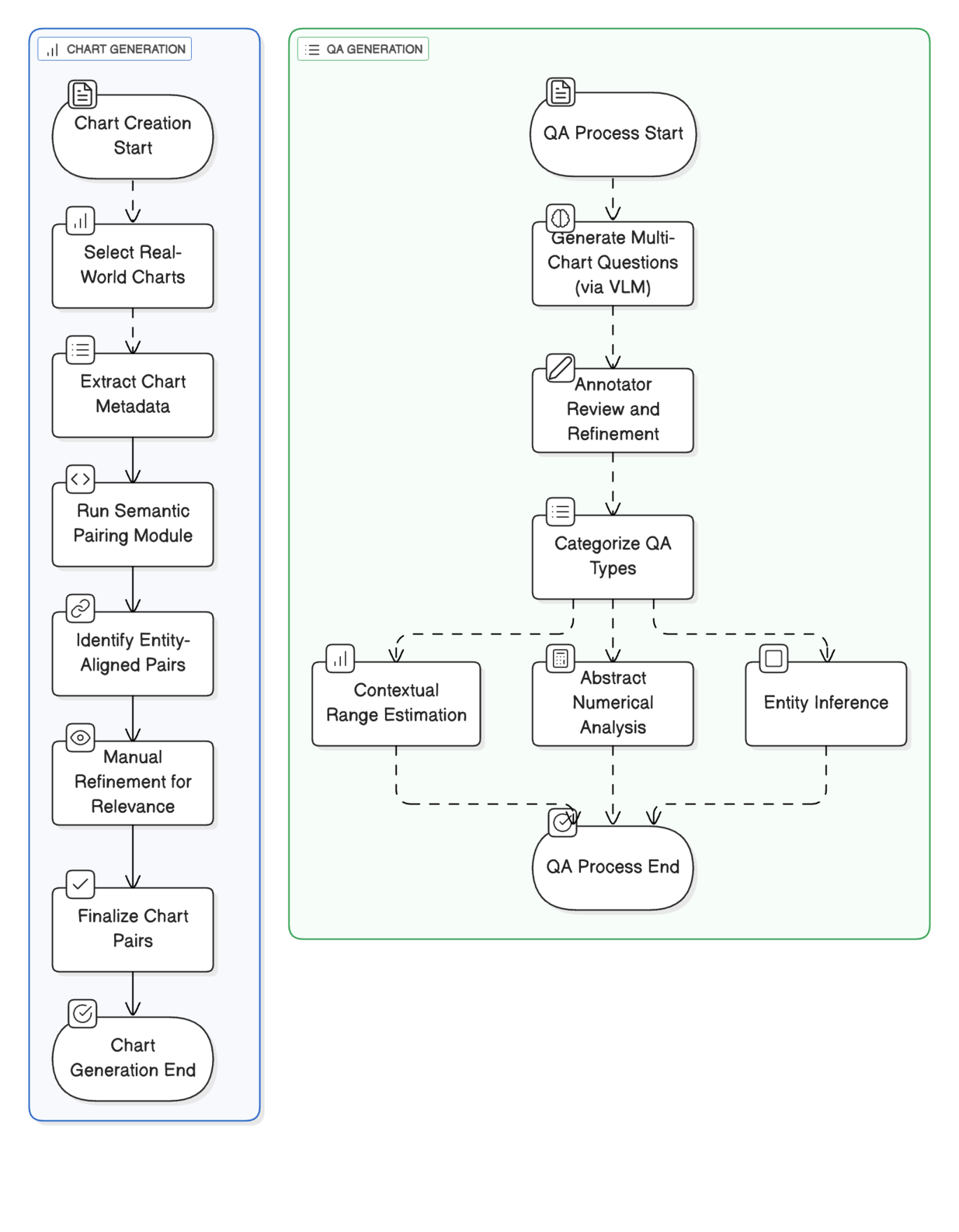}
        \caption{Pipeline for STORM: Constructing real-world chart pairs and QA for multi-step reasoning across misaligned domains.}
        \label{fig:s3}
    \end{minipage}
\end{figure*}

\begin{figure}
    \centering
    \includegraphics[width=0.9\linewidth]{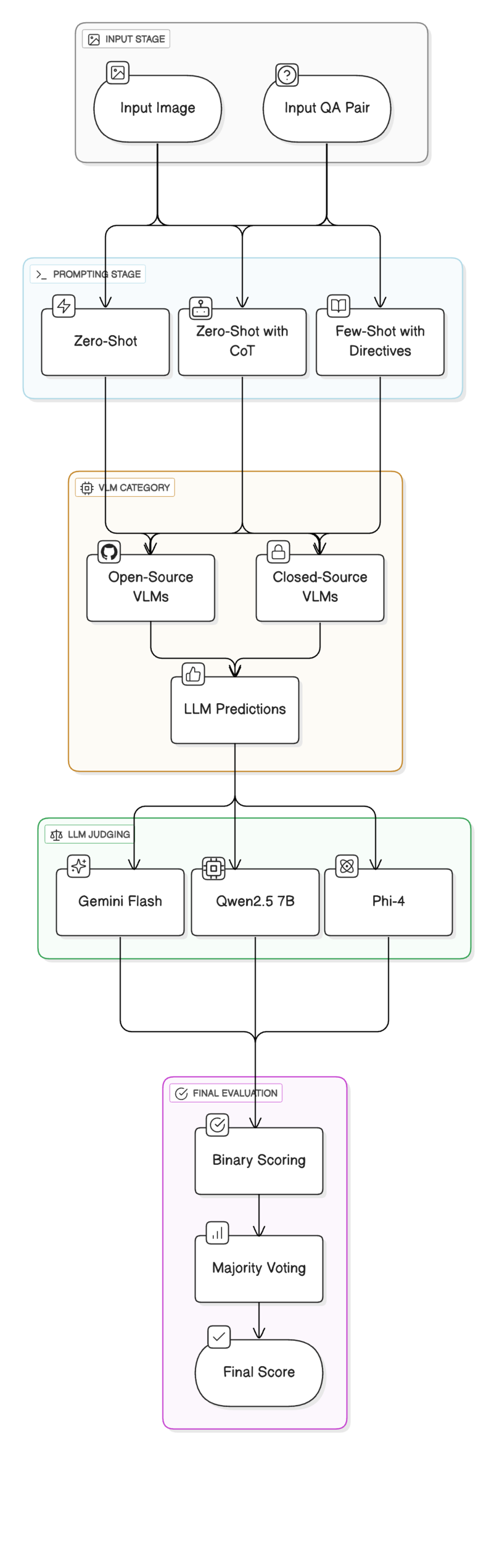}
    \caption{Evaluation pipeline overview: Combining chart-question inputs with different prompting strategies and judging model outputs via majority voting from multiple LLMs.}
    \label{fig:flowchart_eval}
\end{figure}

\newpage
\section{Data Generation Algorithms}
\label{app:data_gene_algo}
\begin{algorithm}
  \caption{\textbf{\textit{DECAF}} Constrained SQL Sampling -Multi-Entity Chart Decomposition}
  \label{alg:sql-sampling}
  \resizebox{0.8\linewidth}{!}{ 
\begin{minipage}{\linewidth}
  \begin{algorithmic}[1]
    \State \textbf{Input:} Table $T$, Level $L$, Operators $OP_{num}$, $OP_{str}$, $FL_{ops}$, $STR_{ops}$, $C_{nj}$
    \State \textbf{Output:} SQL Query $S$
    \For{each column $C$ in $T$}
        \State Identify $C.dataType$
    \EndFor
    \While{not ValidSQL($S$, $T$)}
        \State Initialize empty SQL Query $S$
        \Comment{Chart Decomposition via SQL}
        \State $select\_col \gets$ Random Column from $T$
        \If{$L = 1$ and Random(0,1) = 0}
            \State Skip Selection Operation
        \Else
            \If{$select\_col$ is Numerical}
                \State Apply Numerical Operator
            \Else
                \State Apply String Operator
            \EndIf
        \EndIf
        \Comment{WHERE Clause - Linked Data Points Selection}
        \If{Random(0,1) = 1}
            \State Choose Column $C$, Value $V$, Operator $OP$
            \State Add Condition $C OP V$
        \EndIf
        \Comment{WHERE Clause - Multi-Row and Multi-Column Reasoning}
        \State Extract Numeric Columns
        \State Choose Number of Conditions Based on $L$
        \For{each Condition}
            \State Pick Two Numeric Columns $C_A, C_B$
            \State Add Condition $C_A OP C_B$
        \EndFor
        \Comment{Combine Conditions with Conjunctions for Complex Queries}
        \For{each Condition}
            \State Merge using $C_{nj}$ (AND, OR)
        \EndFor
        \Comment{ORDER BY Clause (For L = 2)}
        \If{$select\_col$ is Numerical and not in Conditions}
            \State Apply ORDER BY with ASC/DESC
        \EndIf
    \EndWhile
    \State \textbf{Filter by Human} \Comment{Ensuring Logical Consistency and Quality}
    \State \Return $S$
  \end{algorithmic}
  \end{minipage}
  }
\end{algorithm}
\begin{algorithm}
  \caption{Synthetic Simulation - Multi-Chart Reasoning with LLM-Generated Contexts}
  \label{alg:synthetic-simulation}
  \begin{algorithmic}[1]
    \State \textbf{Input:} LLM Model $M_{LLM}$, Human Annotators $A$, Chart Generator $G_{chart}$
    \State \textbf{Output:} Dataset $D$ with Context Pairs and QA Pairs

    \Comment{Step 1: Context Table and Chart Generation}
    \State $T_{contexts} \gets \emptyset$
    \For{each scenario $S$ generated by $M_{LLM}$}
        \State Extract structured entity relationships $E_S$
        \State Construct context tables $T_S$ based on $E_S$
        \State $T_{contexts} \gets T_{contexts} \cup T_S$
    \EndFor
    
    \State $C_{synthetic} \gets \emptyset$
    \For{each table $T$ in $T_{contexts}$}
        \State Convert $T$ into chart $C$ using $G_{chart}$
        \State Perform human review for accuracy and readability
        \State $C_{synthetic} \gets C_{synthetic} \cup C$
    \EndFor
    
    \Comment{Step 2: Multi-Chart QA Generation}
    \State $QA \gets \emptyset$
    \For{each related chart pair $(C_1, C_2)$ in $C_{synthetic}$}
        \For{each annotator $a$ in $A$}
            \State Generate Questions
            \State Use LLM-based prompt chaining for QA refinement
        \EndFor
    \EndFor
    
    \Comment{Step 3: Dataset Filtering and Compilation}
    \State Perform Human Validation for Correctness and Clarity
    \State Remove Low-Quality QA Pairs
    \State $D \gets \{C_{synthetic}, QA\}$
    \State \Return $D$
  \end{algorithmic}
\end{algorithm}

\begin{algorithm}[t]
\caption{STORM: Chart and QA Generation}
\label{alg:realm}
\begin{algorithmic}[1]
\State \textbf{Input:} Chart Repository $\mathcal{C}$, Semantic Pairing Module $P_{sem}$, VLM Model $M_{VLM}$, Annotators $A$
\State \textbf{Output:} Dataset $D = \{(C_i, C_j, q, a)\}$

\Statex
\State \hfill \textbf{// Chart Generation Phase}
\State Initialize paired chart set $\mathcal{P}_{final} \gets \emptyset$
\For{each chart $C_i$ in repository $\mathcal{C}$}
    \State Extract metadata $M_{C_i}$
    \State Use $P_{sem}$ to find matching chart $C_j$ with aligned entities
    \If{valid alignment exists}
        \State Add $(C_i, C_j)$ to candidate pairs
    \EndIf
\EndFor
\For{each pair $(C_i, C_j)$ in candidate pairs}
    \State Manually review for relevance and coherence
    \If{pair is contextually valid}
        \State Add to $\mathcal{P}_{final}$
    \EndIf
\EndFor

\Statex
\State \hfill \textbf{// QA Generation Phase}
\State Initialize QA set $\mathcal{Q} \gets \emptyset$
\For{each chart pair $(C_i, C_j)$ in $\mathcal{P}_{final}$}
    \State Generate candidate QA pairs using $M_{VLM}$
    \State Annotators review and refine each $(q, a)$
    \State Classify QA into one of:
    \begin{itemize}
        \item Contextual Range Estimation
        \item Abstract Numerical Analysis
        \item Entity Inference
    \end{itemize}
    \State Add $(C_i, C_j, q, a)$ to $\mathcal{Q}$
\EndFor

\State \Return Final dataset $D \gets \mathcal{Q}$
\end{algorithmic}
\end{algorithm}


\clearpage

\section{Model and Compute Details}

\paragraph{Model Sizes.} We evaluated a mix of closed- and open-source vision-language models (VLMs), as well as structured reasoning baselines:
\begin{itemize}
    \item \textbf{Gemini 1.5 Pro}: ~56B parameters (proprietary, estimate based on public disclosures).
    \item \textbf{GPT-4o-mini}: Parameter size not publicly disclosed.
    \item \textbf{Qwen2-VL-7B-Instruct}: 7B parameters.
    \item \textbf{MiniCPM-V-2\_6}: 2.6B parameters.
    \item \textbf{InternVL-2-8B}: 8B parameters.
    \item \textbf{Idefics3-8B-LLaMA3}: 8B parameters.
    \item \textbf{DePlot (Liu et al., 2023)}: Built on encoder-decoder transformer with tabular rendering; ~400M parameters.
    \item \textbf{Chart-to-Text (Kantharaj et al., 2022)}: Includes rule-based visual parsing + generation via T5 (220M to 3B parameters, depending on version).
\end{itemize}

\paragraph{Compute Infrastructure.} Model inference and evaluation were performed using:
\begin{itemize}
    \item NVIDIA A100, NVIDIA H200 GPUs on a high-memory compute cluster for open-source model inference and table-based prompting.
    \item Google Cloud and OpenAI APIs for Gemini 1.5 Pro and GPT-4o-mini, respectively.
\end{itemize}

\paragraph{Approximate Compute Budget.}
\begin{itemize}
    \item \textbf{Open-source model inference}: $\sim$320 GPU-hours (covering 5,214 QA pairs × 3 prompting strategies × multiple visual formats).
    \item \textbf{Evaluation with LLM-as-a-Judge}: $\sim$60 GPU-hours (Gemini 1.5 Flash, Qwen2.5-7B, and Phi-4; each example judged by 3 models).
    \item \textbf{Chart-to-Table + Table-based QA (DePlot, DePlot++, Gemini, MiniCPM, Qwen2)}: $\sim$120 GPU-hours for rendering, metadata generation, and table-based prompting.
\end{itemize}

All experiments were implemented in Python $\geq$ 3.10 using PyTorch $\geq$ 2.0. Evaluation workflows used batch inference pipelines with structured logging, and charts were rendered or parsed using Plotly, DePlot, and in-house scripts.

\begin{table}[H]
\section{STORM Results: GPT-4o}
\label{tab:gpt4o_appendix}
\small
\centering
\renewcommand{\arraystretch}{1.05}
\setlength{\tabcolsep}{5pt}
\begin{tabular}{lcc}
\toprule
\multirow{2}{*}{\textbf{GPT-4o Performance}} & \multicolumn{2}{c}{\textit{STROM}}  \\
\cmidrule(lr){2-3}
& \textbf{Combined} & \textbf{Interleaved} \\
\midrule
\textbf{Zero-Shot} & 37.1 & 36.7 \\
\textbf{Zero-Shot CoT} & 35.3 & 36.0 \\
\textbf{Directives (Few-Shot CoT$_D$)}& 36.1 & 33.3 \\
\bottomrule
\end{tabular}
\caption{Detailed accuracies for \textbf{GPT-4o} across prompting strategies and visual contexts. These results highlight that GPT-4o’s scores approach Gemini-1.5 Pro, particularly within the \textit{STORM} subset emphasizing semantic and temporal reasoning.}
\end{table}

\subsection*{Comparison Between Gemini-1.5 Pro and GPT-4o on STORM.}
Gemini-1.5 Pro and GPT-4o exhibit comparable trends on the \textit{STORM} subset, which emphasizes semantic abstraction and temporal reasoning. Under the combined visual context, Gemini-1.5 Pro achieves accuracies of 34.8\%, 34.9\%, and 33.7\% across Zero-Shot, Zero-Shot CoT, and Directive prompting, respectively, while GPT-4o attains 37.1\%, 35.3\%, and 36.1\% under the same conditions. In the interleaved format, Gemini-1.5 Pro records 36.0\%, 36.4\%, and 32.9\%, compared to GPT-4o’s 36.7\%, 36.0\%, and 33.3\%. These results indicate that GPT-4o performs on par with Gemini-1.5 Pro within STORM, showing slightly higher stability across prompting strategies and marginally better outcomes under the combined context. Both models, however, display limited gains from chain-of-thought prompting, underscoring the persistent difficulty of multi-chart temporal abstraction even for top-tier proprietary VLMs.

\section{Example of STROM}
\label{fig:storm_example_pair}
The STORM subset draws from real-world chart pairs curated from \emph{Our World in Data}, where authors frequently embed two thematically linked line charts within the same figure. These pairs mirror authentic analytical practices in domains such as public health, economics, and climate reporting. 

\begin{figure}[t]
\centering
\begin{subfigure}[t]{0.47\textwidth}
    \centering
  \includegraphics[height=5cm,width=\linewidth,keepaspectratio]{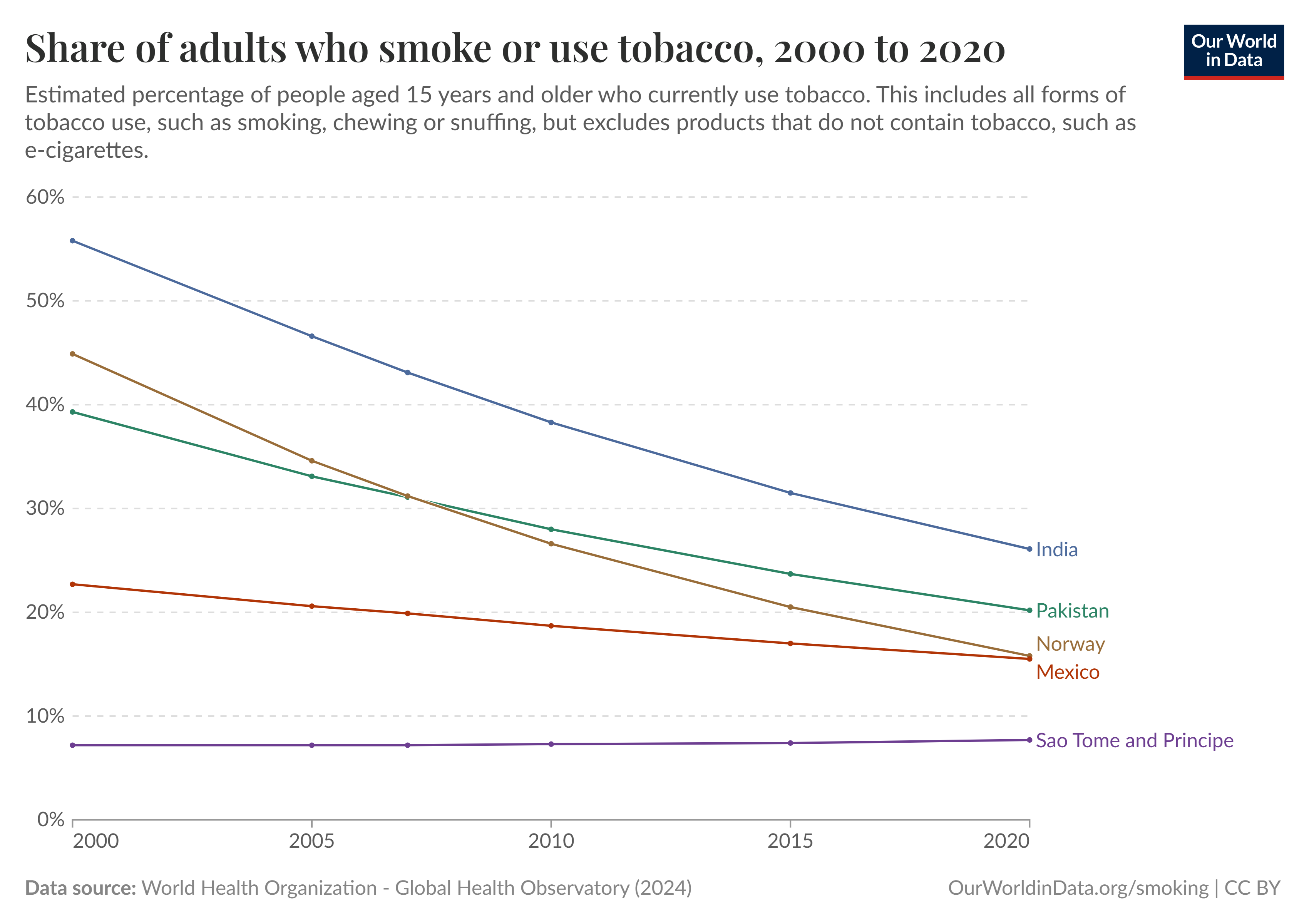}
  \label{fig:storm_q_chart1}
\end{subfigure}
\hfill
\begin{subfigure}[t]{0.47\textwidth}
\centering
  \includegraphics[height=5cm,width=\linewidth,keepaspectratio]{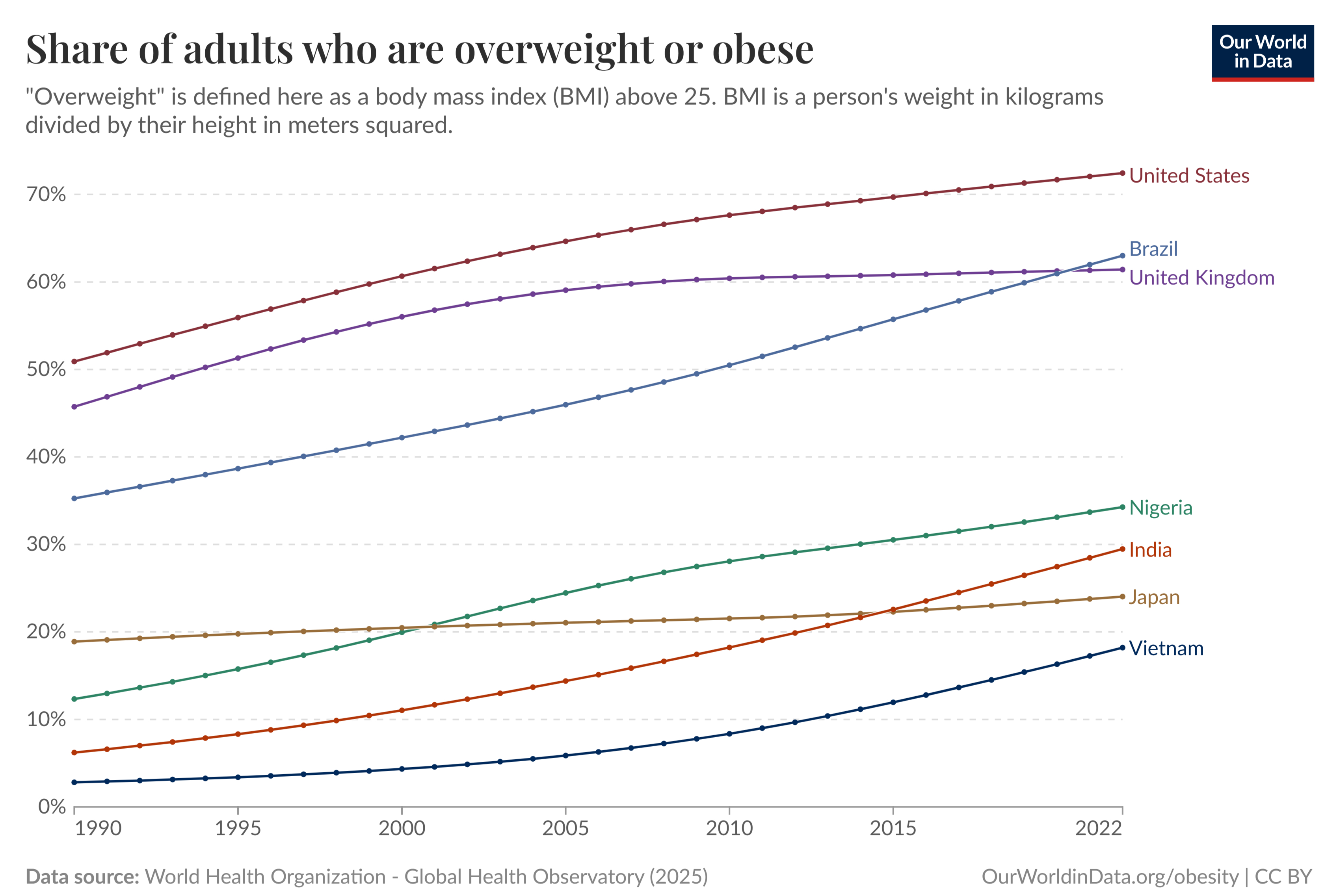}
  \label{fig:storm_q_chart2}
\end{subfigure}

\caption{
\textbf{Question:} During the approximate period from 2015 to 2020, when the share of overweight or obese adults in India was approaching and surpassing that of Japan, what was the corresponding range for the share of adults in Norway who smoke or use tobacco?\\[3pt]
\textbf{Answer:} 15.8-20.5\% \\ 
}
\end{figure}

This question requires cross-chart temporal reasoning: identifying overlapping time windows in two charts, correlating independent metrics, and estimating a numerical range tasks that demand visual alignment and contextual interpretation beyond local fact retrieval.

\begin{figure}[t]
    \centering
    \includegraphics[width=\linewidth]{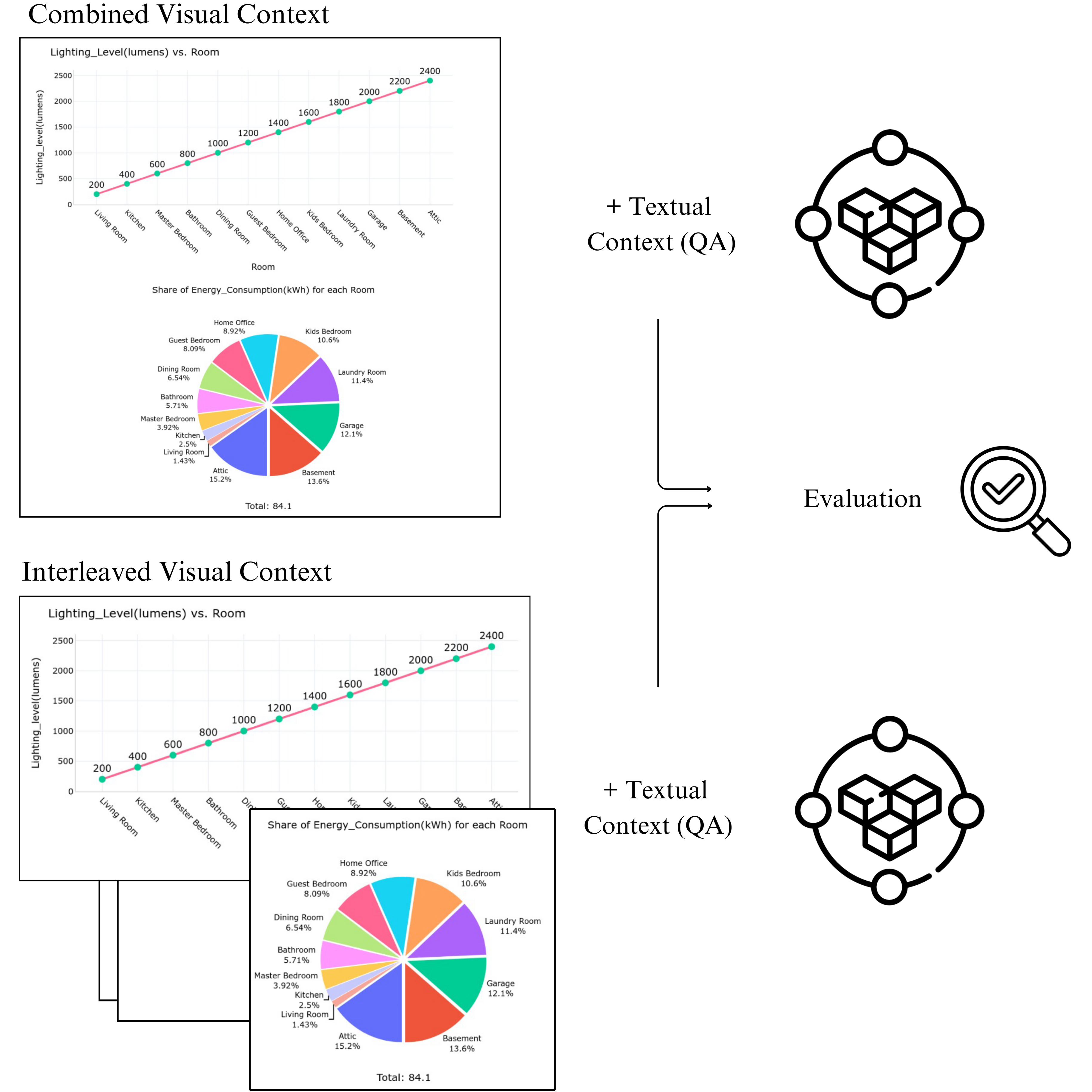}
    \caption{Visual input formats in \InterChart{}: \textbf{Combined} (stitched multi-chart image) vs. \textbf{Interleaved} (separate sequential chart images).}
    \label{fig:visual_input_formats}
\end{figure}

\section{Annotation Instructions}
\label{appendix:annotation_instructions}

To ensure consistency and reliability, all annotators received standardized written guidelines outlining task objectives, labeling rules, and arbitration policies. Below is a condensed version of the instruction set provided during dataset construction.

\paragraph{Objective.}  
Annotators were asked to verify question-answer (QA) pairs for correctness, clarity, and grounding in the associated chart(s). Each QA pair should be answerable directly from the visual and textual content of the chart without requiring external knowledge.

\paragraph{Procedure.}  
1. Read the question carefully and locate all referenced visual elements in the chart (axes, legends, data points, text).  
2. Identify the correct answer span or value directly from the chart.  
3. For numerical answers, record values up to one decimal precision; for textual answers, reproduce the label verbatim.  
4. If a question is ambiguous, inconsistent, or unanswerable, flag it for arbitration rather than guessing.  

\paragraph{Annotation Policy.}  1. Each item was independently annotated by two annotators.  
2. Disagreements were reviewed by a senior annotator following predefined arbitration rules.  
3. Only consensus or majority-agreed entries were retained in the final dataset.  

\paragraph{Examples Provided.}  
Annotators were given representative examples covering: 
(i) single-chart factual QA,  
(ii) trend-based correlation QA, and  
(iii) multi-chart temporal inference QA.  

All instruction documents, example templates, and arbitration notes will be released as part of the supplementary materials for transparency and reproducibility.

\section{Discussion on Evaluation Metrics}
\label{appendix:automatic_metrics}

While this work primarily relies on large language models (LLMs) as semantic judges for evaluating answer correctness, we also considered whether conventional automatic metrics such as BLEURT~\cite{sellam2020bleurt}, MoverScore~\cite{zhao2019moverscore}, and QuestEval~\cite{scialom2021questeval} could be applied to vision-language model (VLM) evaluation.

\paragraph{Limitations of Token- and Embedding-Based Metrics.}
Metrics like BLEURT and MoverScore depend on lexical or embedding-level similarity between the generated and reference answers. However, multi-chart visual question answering often involves paraphrased numerical expressions (e.g., ``around 25\%'' vs. ``roughly one-fourth''), varied unit conversions, or inferred temporal relations that cannot be captured reliably through surface-level similarity. These metrics thus underestimate correctness when models provide semantically valid but lexically diverse responses.

\paragraph{Advantages of LLM-Based Semantic Evaluation.}
Our LLM-judge framework allows contextual reasoning and flexible comparison through majority voting across multiple evaluators. This setup accounts for approximate numeric equivalence, synonymous phrasing, and task-specific conditions such as tolerance for rounding or unit mismatches. It better aligns with human judgment for multi-step reasoning tasks involving quantitative and relational inference.

\paragraph{Complementary Role of Automatic Metrics.}
Although limited, BLEURT, MoverScore, and QuestEval can still serve as lightweight indicators of linguistic fidelity or surface coherence, particularly in benchmarking language fluency rather than reasoning correctness. In future work, integrating these metrics alongside semantic judges could yield a hybrid evaluation pipeline combining automatic reproducibility with reasoning-aware assessment to improve the interpretability and comparability of VLM performance.

\section{Benchmark Comparison}
\label{appendix:comparison}
As summarized in Table~\ref{tab:benchmark_comparison}, \InterChart{} complements recent benchmarks such as InfoChartQA and ChartQAPro by explicitly introducing controlled multi-chart reasoning under real-world visual conditions. While InfoChartQA and ChartMind feature high visual diversity through infographics and mixed formats, their tasks primarily test factual or single-chart reasoning without requiring temporal alignment or semantic aggregation across figures. ChartQAPro focuses on single-plot factual reasoning with limited variation in visual structure, making it less diagnostic of multi-source reasoning failures. In contrast, \InterChart{} isolates specific sources of difficulty semantic drift, temporal reasoning, and cross-chart integration through its tiered subsets (DECAF, SPECTRA, and STORM). This layered design enables ablation-style analysis of reasoning failure modes: models that perform well on single-chart subsets (e.g., DECAF) often degrade sharply on STORM, revealing deficits not in visual extraction but in temporal synthesis and semantic correlation. \InterChart{} complements existing single-chart datasets by serving as a diagnostic benchmark rather than a direct performance leaderboard.

\section{Individual Evaluation Results}
\label{app:appendix_d}
Tables~\ref{tab:phi_results_placeholder}, \ref{tab:gemini_results}, and \ref{tab:qwen_results_placeholder} present accuracy distributions when using \textbf{Phi-4}, \textbf{Gemini-1.5-Pro}, and \textbf{Qwen-2.5-Instruct} as independent semantic judges. 
Across all evaluators, trends remain consistent \textit{DECAF} achieves the highest accuracy, followed by \textit{SPECTRA}, with a pronounced decline on \textit{STORM}, confirming the benchmark’s difficulty gradient. 
However, absolute scores vary by evaluator due to differences in model size, reasoning strength, and sensitivity to paraphrased answers.

Among the three, \textbf{Gemini-1.5-Pro} yields the most lenient yet semantically consistent judgments, particularly on \textit{STORM}, where its accuracy remains higher for both zero-shot and CoT strategies. 
\textbf{Phi-4} produces moderate scores with stable cross-tier variance, reflecting balanced precision and recall for numerical and textual reasoning tasks. 
In contrast, \textbf{Qwen-2.5-Instruct} exhibits a stricter interpretation of correctness, often penalizing minor phrasing or unit mismatches leading to slightly lower absolute accuracies but higher alignment with human annotation consistency observed in Section~\ref{sec:evaluation}.

Together, these results demonstrate that the multi-judge setup captures complementary evaluation perspectives: Gemini offers semantic flexibility, Phi-4 ensures numerical reliability, and Qwen enforces syntactic rigor. 
This diversity supports the robustness of our majority-voting framework, where aggregate correctness judgments remain stable despite evaluator-level variability.

\clearpage
\phantomsection

\begin{table*}[t]
\centering
\captionsetup{width=\textwidth,justification=justified,singlelinecheck=false}
\small
\renewcommand{\arraystretch}{1.2}
\setlength{\tabcolsep}{14pt}
\begin{tabular}{l|c|c|c|c}
\toprule
\textbf{Dimension} & \textbf{InfoChartQA} & \textbf{ChartMind} & \textbf{ChartQAPro} & \textbf{\InterChart{}} \\
\toprule
\toprule
\textbf{Chart Type} & Infographics & Mixed & Plots & Plots \\
\midrule
\textbf{Multi-Chart} & No & Limited & No & \textbf{Yes} \\
\midrule
\textbf{Real-World Data} & Yes & Yes & Yes & \textbf{Yes} \\
\midrule
\textbf{Semantic Drift} & Medium & Medium & Low & \textbf{High} \\
\midrule
\textbf{Temporal Reasoning} & Low & Medium & Low & \textbf{High} \\
\midrule
\textbf{Visual Diversity} & High & High & Low & \textbf{High} \\
\midrule
\textbf{QA Type} & Factoid & Hybrid & Factual & Fact + Inference \\
\midrule
\textbf{Evaluation Method} & BLEURT & BLEU / LLM & Exact Match & \textbf{LLM Majority Voting} \\
\bottomrule
\end{tabular}
\caption{Comparison of recent benchmarks across diagnostic dimensions. 
\InterChart{} uniquely supports rigorous multi-chart reasoning with high semantic and temporal complexity, particularly through its \emph{STORM} subset.}
\label{tab:benchmark_comparison}
\end{table*}

\begin{table*}[!htbp]
\centering
\renewcommand{\arraystretch}{1.5}
\small
\setlength{\tabcolsep}{3pt}
\begin{tabular}{lcccccccccccc}
  \toprule
  \textbf{Model} & \multicolumn{4}{c}{\textbf{Zero-Shot}} & \multicolumn{4}{c}{\textbf{Zero-Shot CoT}} & \multicolumn{4}{c}{\textbf{Few-Shot CoT\textsubscript{D}}} \\
& \textbf{\footnotesize Net} & \textbf{\scriptsize DECAF} & \textbf{\scriptsize SPECTRA} & \textbf{\scriptsize STORM} & \textbf{\footnotesize Net} & \textbf{\scriptsize DECAF} & \textbf{\scriptsize SPECTRA} & \textbf{\scriptsize STORM} & \textbf{\footnotesize Net} & \textbf{\scriptsize DECAF} & \textbf{\scriptsize SPECTRA} & \textbf{\scriptsize STORM} \\
  \midrule
  \multicolumn{13}{c}{\textit{Combined Visual Context Image}} \\
  \midrule
  GPT-4o-mini              & 47.5 & 61.8 & 49.4 & 31.3 & 53.3 & 73.8 & 55.8 & 30.5 & 53.0 & 74.8 & 53.2 & 31.0 \\
  Gemini-1.5-Pro     & 53.0 & 67.1 & 61.1 & 30.9 & 54.6 & 73.9 & 58.6 & 31.4 & 57.8 & 78.4 & 64.4 & 30.5 \\
  Qwen2-VL-7B           & 38.3 & 52.2 & 35.7 & 27.0 & 46.0 & 66.5 & 45.4 & 27.0 & 44.2 & 63.1 & 46.8 & 22.8 \\
  MiniCPM-V-2\_6            & 38.9 & 57.9 & 37.6 & 21.3 & 38.6 & 58.7 & 37.2 & 20.0 & 36.6 & 50.2 & 39.0 & 20.6 \\
  InternVL-2-8B           & 34.1 & 44.6 & 32.5 & 25.0 & 37.9 & 53.6 & 36.6 & 23.7 & 36.9 & 53.4 & 35.5 & 21.9 \\
  Idefics3-8B-Llama3            & 27.7 & 41.6 & 19.9 & 21.8 & 28.9 & 40.6 & 21.4 & 24.6 & 27.7 & 38.6 & 27.8 & 16.7 \\
  \midrule
  \textbf{Mean}      & 39.9 & 54.2 & 39.4 & 26.0 & 43.2 & 61.2 & 42.5 & 26.2 & 42.0 & 59.8 & 44.4 & 23.9 \\
  \midrule
  \multicolumn{13}{c}{\textit{Interleaved Visual Context}} \\
  \midrule
  GPT-4o-mini               & 55.1 & 68.5 & 53.6 & 33.8 & 56.0 & 77.6 & 56.6 & 33.5 & 55.8 & 77.7 & 55.6 & 34.1 \\
  Gemini-1.5-Pro     & 55.3 & 74.5 & 58.4 & 33.1 & 55.9 & 76.4 & 57.9 & 33.4 & 57.1 & 78.0 & 63.5 & 29.9 \\
  Qwen2-VL-7B             & 37.3 & 49.4 & 35.2 & 27.3 & 45.9 & 64.6 & 44.6 & 28.6 & 42.0 & 55.3 & 44.3 & 26.4 \\
  MiniCPM-V-2\_6            & 45.0 & 66.0 & 44.0 & 25.0 & 43.4 & 64.1 & 42.0 & 24.2 & 44.0 & 63.3 & 44.4 & 24.3 \\
  \midrule
  \textbf{Mean}      & 48.2 & 64.6 & 47.8 & 29.8 & 50.3 & 70.7 & 50.3 & 29.9 & 49.7 & 68.6 & 51.9 & 28.7 \\
\bottomrule
\end{tabular}

\caption{Baseline Accuracies using our evaluation method with Microsoft Phi4 Eval Engine on All Models and Strategies, broken down by Set Type (S1, S2, S3) and Strategy type (Zero-Shot, Zero-Shot CoT, Few-Shot CoT$_D$).}
\label{tab:phi_results_placeholder}
\end{table*}
\clearpage
\phantomsection

\twocolumn[{%

  \begin{center}
  \captionof{table}{Baseline Accuracies using our evaluation method with Gemini-1.5 Eval Engine on all models and prompting strategies. Results are grouped by visual context format (top: Combined, bottom: Interleaved), and broken down by set type (DECAF, SPECTRA, STORM) and strategy (Zero-Shot, Zero-Shot CoT, Few-Shot CoT$_D$).\\}
  \label{tab:gemini_results}
  \small
  \renewcommand{\arraystretch}{1.4}
  \setlength{\tabcolsep}{3pt}
  \begin{tabular}{lcccccccccccc}
  \toprule
  \textbf{Model} & \multicolumn{4}{c}{\textbf{Zero-Shot}} & \multicolumn{4}{c}{\textbf{Zero-Shot CoT}} & \multicolumn{4}{c}{\textbf{Few-Shot CoT\textsubscript{D}}} \\
& \textbf{\footnotesize Net} & \textbf{\scriptsize DECAF} & \textbf{\scriptsize SPECTRA} & \textbf{\scriptsize STORM} & \textbf{\footnotesize Net} & \textbf{\scriptsize DECAF} & \textbf{\scriptsize SPECTRA} & \textbf{\scriptsize STORM} & \textbf{\footnotesize Net} & \textbf{\scriptsize DECAF} & \textbf{\scriptsize SPECTRA} & \textbf{\scriptsize STORM} \\
  \midrule
  \multicolumn{13}{c}{\textit{Combined Visual Context Image}} \\
  \midrule
  GPT-4o-mini               & 45.8 & 60.9 & 48.5 & 27.9 & 48.0 & 69.8 & 47.2 & 27.1 & 48.0 & 69.4 & 45.5 & 29.0 \\
  Gemini-1.5-Pro      & 56.3 & 66.3 & 61.7 & 40.8 & 59.3 & 73.8 & 62.0 & 42.2 & 59.1 & 74.6 & 62.9 & 39.9 \\
  Qwen2-VL-7B         & 48.7 & 50.3 & 33.8 & 35.2 & 51.0 & 60.7 & 36.6 & 33.9 & 47.8 & 55.6 & 34.5 & 33.3 \\
  MiniCPM-V-2\_6      & 38.0 & 53.4 & 34.0 & 26.5 & 38.4 & 53.9 & 33.5 & 27.8 & 33.5 & 50.8 & 27.7 & 22.1 \\
  InternVL-2-8B       & 33.2 & 40.3 & 27.8 & 31.6 & 31.6 & 43.4 & 26.2 & 28.6 & 31.4 & 44.3 & 22.4 & 27.6 \\
  Idefics3-8B-Llama3  & 22.2 & 38.2 & 19.6 & 8.9  & 23.0 & 38.1 & 18.3 & 12.8 & 29.0 & 33.5 & 27.0 & 26.6 \\
  \midrule
  \textbf{Mean}       & 40.7 & 51.6 & 37.6 & 28.2 & 42.2 & 56.6 & 37.3 & 28.9 & 41.5 & 54.7 & 36.7 & 29.8 \\
  \midrule
  \multicolumn{13}{c}{\textit{Interleaved Visual Context}} \\
  \midrule
  GPT-4o-mini               & 49.3 & 66.1 & 52.2 & 29.7 & 51.8 & 74.0 & 50.9 & 30.6 & 50.6 & 73.0 & 49.8 & 29.0 \\
  Gemini-1.5-Pro      & 59.0 & 74.2 & 62.9 & 43.0 & 60.0 & 75.0 & 61.9 & 43.0 & 58.4 & 76.1 & 61.3 & 39.4 \\
  Qwen2-VL-7B         & 47.5 & 47.6 & 34.1 & 30.8 & 50.3 & 59.6 & 38.8 & 32.5 & 45.1 & 52.5 & 32.5 & 30.2 \\
  MiniCPM-V-2\_6      & 41.7 & 59.1 & 36.6 & 29.3 & 41.0 & 57.1 & 37.2 & 28.9 & 38.2 & 53.3 & 32.2 & 29.1 \\
  \midrule
  \textbf{Mean}       & 49.4 & 61.7 & 46.5 & 33.2 & 50.8 & 66.4 & 47.2 & 33.8 & 48.1 & 63.7 & 43.9 & 31.9 \\
  \bottomrule
\end{tabular}

  \end{center}

  \vspace{1.5em}

  \begin{center}
  \captionof{table}{Baseline Accuracies using our evaluation method with Qwen 2.5 Eval Engine on all models and prompting strategies. Results are grouped by visual context format (top: Combined, bottom: Interleaved), and broken down by set type (S1, S2, S3) and strategy (Zero-Shot, Zero-Shot CoT, Few-Shot CoT$_D$).\\}
  \label{tab:qwen_results_placeholder}
  \small
  \renewcommand{\arraystretch}{1.4}
  \setlength{\tabcolsep}{3pt}
  \begin{tabular}{lcccccccccccc}
  \toprule
  \textbf{Model} & \multicolumn{4}{c}{\textbf{Zero-Shot}} & \multicolumn{4}{c}{\textbf{Zero-Shot CoT}} & \multicolumn{4}{c}{\textbf{Few-Shot CoT\textsubscript{D}}} \\
  & \footnotesize Net & \scriptsize DECAF & \scriptsize SPECTRA & \scriptsize STORM & \footnotesize Net & \scriptsize DECAF & \scriptsize SPECTRA & \scriptsize STORM & \footnotesize Net & \scriptsize DECAF & \scriptsize SPECTRA & \scriptsize STORM \\
  \midrule
  \multicolumn{13}{c}{\textit{Combined Visual Context Image}} \\
  \midrule
  GPT-4o-mini           & 41.4 & 55.3 & 38.8 & 30.1 & 44.2 & 61.2 & 40.8 & 30.6 & 45.2 & 61.7 & 42.8 & 31.1 \\
  Gemini-1.5-Pro        & 51.1 & 66.1 & 54.5 & 32.6 & 51.1 & 67.0 & 54.9 & 31.4 & 52.1 & 68.6 & 57.1 & 30.7 \\
  Qwen2-VL-7B           & 33.8 & 48.0 & 29.0 & 24.5 & 35.5 & 52.5 & 29.8 & 24.3 & 34.5 & 50.1 & 29.8 & 23.7 \\
  MiniCPM-V-2\_6         & 29.1 & 45.2 & 25.6 & 16.6 & 28.9 & 45.4 & 25.2 & 16.2 & 27.3 & 45.2 & 23.4 & 13.2 \\
  InternVL-2-8B         & 24.3 & 35.1 & 19.6 & 18.2 & 26.3 & 38.6 & 21.8 & 18.5 & 26.6 & 41.4 & 24.1 & 14.2 \\
  Idefics3-8B-Llama3    & 19.8 & 38.1 & 18.8 & 2.5  & 19.5 & 37.7 & 18.9 & 2.0  & 19.7 & 34.9 & 20.4 & 3.9  \\
  \textbf{Mean}         & 33.2 & 48.0 & 31.1 & 20.8 & 34.6 & 50.4 & 31.9 & 20.5 & 34.2 & 50.3 & 32.9 & 19.5 \\
  \midrule
  \multicolumn{13}{c}{\textit{Interleaved Visual Context}} \\
  \midrule
  GPT-4o-mini           & 45.6 & 61.3 & 44.1 & 31.4 & 47.3 & 65.8 & 44.3 & 31.8 & 48.0 & 65.6 & 47.2 & 31.1 \\
  Gemini-1.5-Pro        & 50.0 & 67.0 & 51.0 & 31.9 & 51.6 & 68.1 & 53.8 & 32.9 & 51.3 & 70.3 & 54.1 & 29.5 \\
  Qwen2-VL-7B           & 33.5 & 46.3 & 29.5 & 24.7 & 36.4 & 51.4 & 32.6 & 25.1 & 34.1 & 48.7 & 28.8 & 24.7 \\
  MiniCPM-V-2\_6         & 34.4 & 52.3 & 29.6 & 21.3 & 32.9 & 49.9 & 29.4 & 19.4 & 32.2 & 49.8 & 28.7 & 18.2 \\
  \textbf{Mean}         & 40.9 & 56.7 & 38.6 & 27.3 & 42.1 & 58.8 & 40.0 & 27.3 & 41.4 & 58.6 & 39.7 & 25.9 \\
  \bottomrule
  \end{tabular}
  \end{center}
}]

\end{document}